\documentclass{article}
\usepackage{footnote}
\makesavenoteenv{figure}
\usepackage{arxiv}
\usepackage[utf8]{inputenc}
\usepackage[T1]{fontenc}
\usepackage{CJKutf8}
\usepackage{mathptmx}
\usepackage{times}
\usepackage{makecell}

\usepackage[dvipsnames]{xcolor}
\definecolor{kimiblue}{rgb}{0.09,0.5,0.99}
\usepackage[breaklinks,colorlinks,allcolors=kimiblue]{hyperref}
\usepackage{url}            
\usepackage{booktabs}       
\usepackage{amsfonts}       
\usepackage[normalem]{ulem}
\usepackage{xcolor}         
\usepackage{nicefrac}       
\usepackage{textgreek}
\usepackage{graphicx}
\usepackage{listings}
\usepackage[utf8]{inputenc}
\usepackage{doi}
\usepackage{algorithm}
\usepackage{algpseudocode}
\usepackage{multirow}
\usepackage{multicol}
\usepackage{xcolor}
\usepackage{subcaption}
\usepackage[normalem]{ulem}
\usepackage{amsmath,amsfonts,bm}
\usepackage{amssymb,amsthm}

\usepackage{enumitem}
\usepackage{subcaption}
\usepackage{adjustbox}
\usepackage{array}
\usepackage{makecell}
\usepackage{wrapfig}
\usepackage{ulem}
\usepackage[
    backend=biber,
    citestyle=numeric,
    bibstyle=numeric,
    mincitenames=1,
    maxcitenames=1,
    maxbibnames=3,
]{biblatex}
\newcommand{\citep}[1]{\parencite{#1}}
\setlist[itemize,1]{leftmargin=\dimexpr 18pt}
\setlist[enumerate,1]{leftmargin=\dimexpr 18pt}

\title{
\raisebox{-0.15\height}{\includegraphics[width=0.032\textwidth]{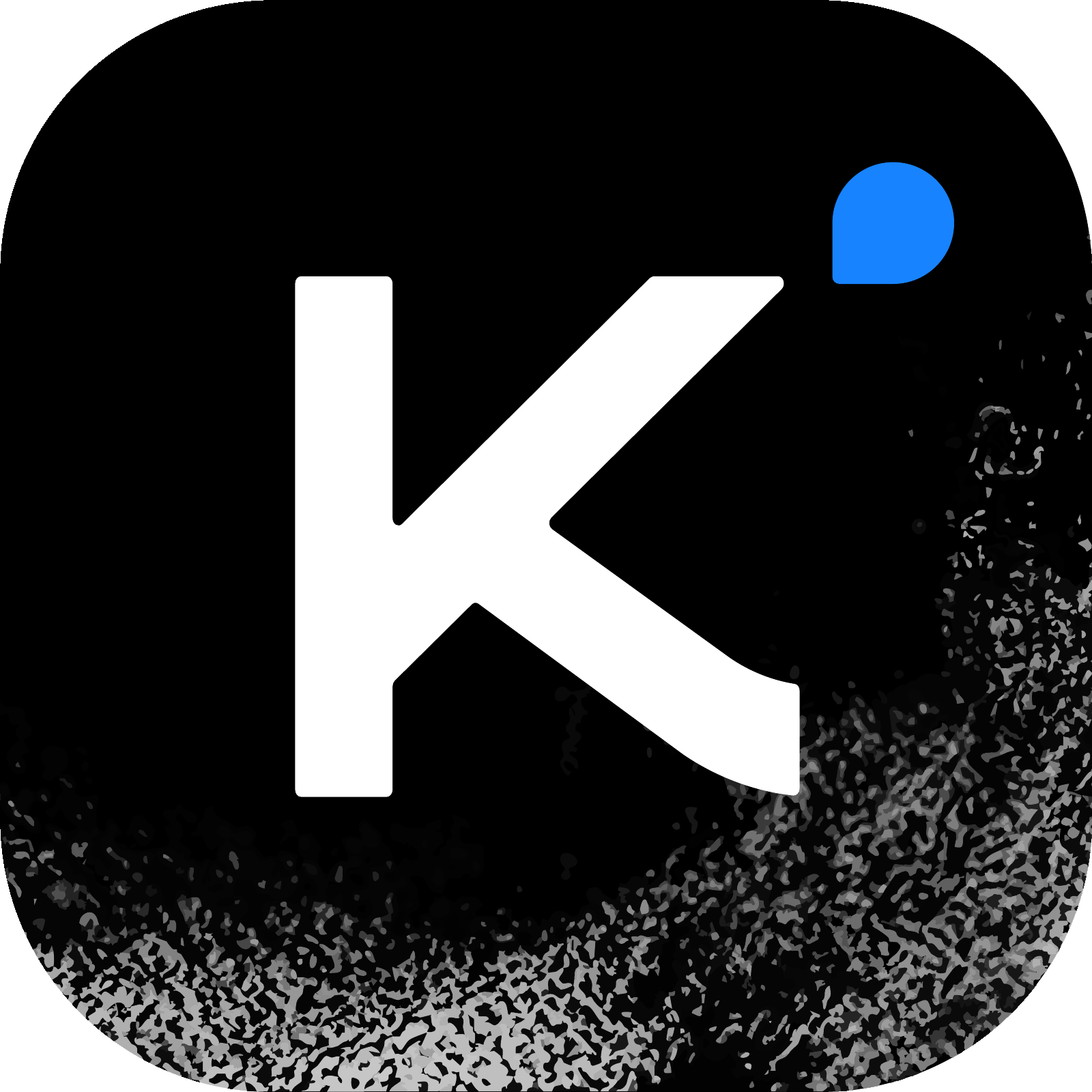}} %
Kimi K2.5: Visual Agentic Intelligence}

\author{Kimi Team}

\date{}

\renewcommand{\headeright}{Technical Report}
\renewcommand{\undertitle}{Technical Report of Kimi K2.5}
\renewcommand{\shorttitle}{\raisebox{-0.12\height}{\includegraphics[width=0.02\textwidth]{figures/logo.pdf}}
Kimi K2.5}

\definecolor{darkblue}{rgb}{0.0, 0.0, 0.5}
\definecolor{darkgreen}{rgb}{0.0, 0.5, 0.0}
\definecolor{darkred}{rgb}{0.5, 0.0, 0.0}
\definecolor{darkpurple}{rgb}{0.5, 0.0, 0.5}

\newcommand{\chinese}[1]{\begin{CJK*}{UTF8}{gbsn}{#1}\end{CJK*}}

\begin{document}
\maketitle

\vspace{-12pt}
\begin{abstract}
We introduce Kimi K2.5, an open-source multimodal agentic model designed to advance general agentic intelligence.
K2.5 emphasizes the joint optimization of text and vision so that two modalities enhance each other. This includes a series of techniques such as joint text-vision pre-training,  zero-vision SFT, and joint text-vision reinforcement learning.
Building on this multimodal foundation, K2.5 introduces Agent Swarm, a self-directed parallel agent orchestration framework that dynamically decomposes complex tasks into heterogeneous sub-problems and executes them concurrently. 
Extensive evaluations show that Kimi K2.5 achieves state-of-the-art results across various domains including coding, vision, reasoning, and agentic tasks. Agent Swarm also reduces latency by up to $4.5\times$  over single-agent baselines. We release the post-trained Kimi K2.5 model checkpoint\footnote{\url{https://huggingface.co/moonshotai/Kimi-K2.5}} to facilitate future research and real-world applications of agentic intelligence. 
\end{abstract}

\begin{figure}[htb]
    \centering
    \includegraphics[width=0.9\textwidth]{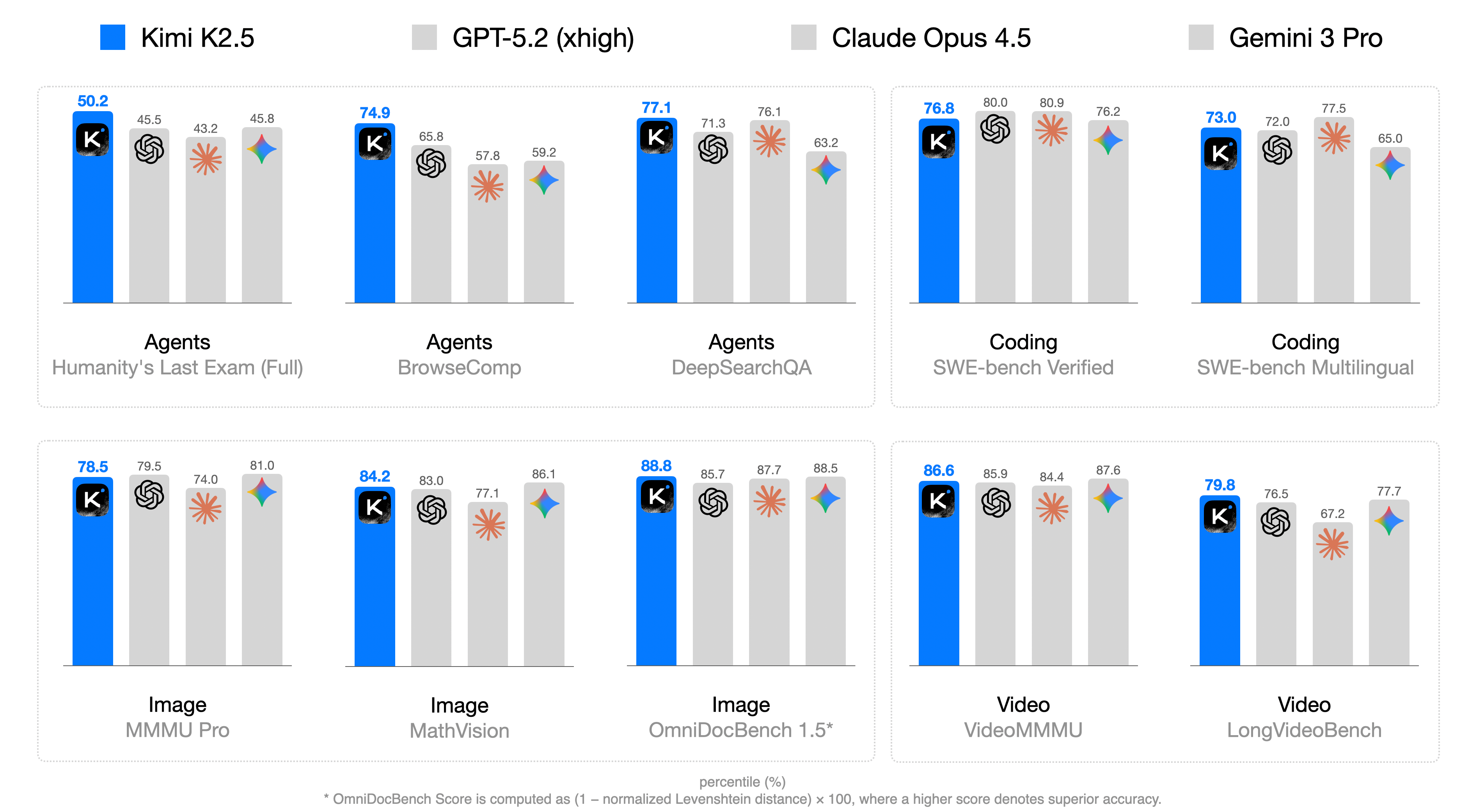}
    \caption{Kimi K2.5 main results.\protect} 
    \label{fig:kimi-k2.5-results}
\end{figure}

\section{Introduction}

Large Language Models (LLMs) are rapidly evolving toward agentic intelligence. Recent advances, such as GPT-5.2~\citep{gpt52report}, Claude Opus 4.5~\citep{opus45report}, Gemini 3 Pro~\citep{gemini3pro}, and Kimi K2-Thinking~\citep{moonshotai2025kimi}, demonstrate substantial progress in agentic capabilities, particularly in tool calling and reasoning. These models increasingly exhibit the ability to decompose complex problems into multi-step plans and to execute long sequences of interleaved reasoning and actions.

In this report, we introduce the training methods and evaluation results of Kimi K2.5. Concretely, we improve the training of K2.5 over previous models in the following two key aspects.

\textbf{Joint Optimization of Text and Vision.} A key insight from the practice of K2.5 is that joint optimization of text and vision enhances both modalities and avoids the conflict. Specifically, we devise a set of techniques for this purpose. During pre-training, in contrast to conventional approaches that add visual tokens to a text backbone at a late stage~\citep{bai2025qwen3vltechnicalreport,guo2025seed15vltechnicalreport}, we find early vision fusion with lower ratios tends to yield better results given the fixed total vision-text tokens. Therefore, K2.5 mixes text and vision tokens with a constant ratio throughout the entire training process.

Architecturally, Kimi K2.5 employs MoonViT-3D, a native-resolution vision encoder incorporating the NaViT packing strategy~\citep{dehghani2023patchnpacknavit}, enabling variable-resolution image inputs. For video understanding, we introduce a lightweight 3D ViT compression mechanism: consecutive frames are grouped in fours, processed through the shared MoonViT encoder, and temporally averaged at the patch level. This design allows Kimi K2.5 to process videos up to 4 $\times$ longer within the same context window while maintaining complete weight sharing between image and video encoders.

During post-training, we introduce zero-vision SFT---text-only SFT alone activates visual reasoning and tool use. 
We find that adding human-designed visual trajectories at this stage hurts generalization. 
In contrast, text-only SFT performs better—likely because joint pretraining already establishes strong vision-text alignment, enabling capabilities to generalize naturally across modalities.
We then apply joint RL on both text and vision tasks. 
Crucially, we find visual RL enhances textual performance rather than degrading it, with improvements on MMLU-Pro and GPQA-Diamond. This bidirectional enhancement—text bootstraps vision, vision refines text—represents superior cross-modal alignment in joint training.

\textbf{Agent Swarm: Parallel Agent Orchestration.}
Most existing agentic models rely on sequential execution of tool calls. Even systems capable of hundreds of reasoning steps, such as Kimi K2-Thinking~\citep{moonshotai2025kimi}, suffer from linear scaling of inference time, leading to unacceptable latency and limiting task complexity. As agentic workloads grow in scope and heterogeneity—e.g., building a complex project that involves massive-scale research, design, and development—the sequential paradigm becomes increasingly inefficient.

To overcome the latency and scalability limits of sequential agent execution, Kimi K2.5 introduces \textit{Agent Swarm}, a dynamic framework for parallel agent orchestration. We propose a Parallel-Agent Reinforcement Learning (PARL) paradigm that departs from traditional agentic RL~\citep{moonshotai2025kimiresearcher}. In addition to optimizing tool execution via verifiable rewards, the model is equipped with interfaces for sub-agent creation and task delegation.  During training, sub-agents are frozen and their execution trajectories are excluded from the optimization objective; only the orchestrator is updated via reinforcement learning. This decoupling circumvents two challenges of end-to-end co-optimization: credit assignment ambiguity and training instability. Agent Swarm enables complex tasks to be decomposed into heterogeneous sub-problems executed concurrently by domain-specialized agents, transforming task complexity from linear scaling to parallel processing. In wide-search scenarios, Agent Swarm reduces inference latency by up to 4.5$\times$ while improving item-level F1 from 72.8\% to 79.0\% compared to single-agent baselines.

Kimi K2.5 represents a unified architecture for general-purpose agentic intelligence, integrating vision and language, thinking and instant modes, chats and agents. It achieves strong performance across a broad range of agentic and frontier benchmarks, including state-of-the-art results in visual-to-code generation (image/video-to-code) and real-world software engineering in our internal evaluations, while scaling both the diversity of specialized agents and the degree of parallelism. To accelerate community progress toward General Agentic Intelligence, we open-source our post-trained checkpoints of Kimi K2.5, enabling researchers and developers to explore, refine, and deploy scalable agentic intelligence.

\section{Joint Optimization of Text and Vision}

Kimi K2.5 is a native multimodal model built upon Kimi K2 through large-scale joint pre-training on approximately 15 trillion mixed visual and text tokens. Unlike vision-adapted models that compromise either linguistic or visual capabilities, our joint pre-training paradigm enhances both modalities simultaneously. This section describes the multimodal joint optimization methodology that extends Kimi K2 to Kimi K2.5.

\subsection{Native Multimodal Pre-Training} \label{sec:joint-pre}

\begin{table}[h]
\centering
\caption{Performance comparison across different vision-text joint-training strategies. Early fusion with a lower vision ratio yields better results given a fixed total vision-text token budget.}
\begin{tabular}{lcc|cccccc}
\toprule
 & \shortstack{Vision Injection\\Timing} & \shortstack{Vision-Text\\Ratio} & \shortstack{Vision\\Knowledge} & \shortstack{Vision\\Reasoning} & \shortstack{OCR\\~} & \shortstack{Text\\Knowledge} & \shortstack{Text\\Reasoning} & \shortstack{Code\\~} \\
\midrule
Early & 0\% & 10\%:90\% & \textbf{25.8} & \textbf{43.8} & \textbf{65.7} & \textbf{45.5} & 58.5 & \textbf{24.8} \\
Mid   & 50\% & 20\%:80\% & 25.0 & 40.7 & 64.1 & 43.9 & \textbf{58.6} & 24.0 \\
Late  & 80\% & 50\%:50\% & 24.2 & 39.0 & 61.5 & 43.1 & 57.8 & 24.0 \\
\bottomrule
\end{tabular}
\label{tab:joint-train}
\end{table}

A key design question for multimodal pre-training is: Given a fixed vision-text token budget, what is the optimal vision-text joint-training strategy. Conventional wisdom~\citep{bai2025qwen3vltechnicalreport,guo2025seed15vltechnicalreport} suggests introducing vision tokens predominantly in the later stages of LLM training at high ratios (e.g., 50\% or higher) should accelerate multimodal capability acquisition, treating multimodal capability as a post-hoc add-on to linguistic competence.

However, our experiments (as shown in Table~\ref{tab:joint-train} Figure~\ref{fig:joint-train}) reveal a different story. We conducted ablation studies varying the vision ratio and vision injection timing while keeping the total vision and text token budgets fixed. To strictly meet the targets for different ratios, we pre-trained the model with text-only tokens for a specifically calculated number of tokens before introducing vision data. Surprisingly, we found that the vision ratio has minimal impact on final multimodal performance. In fact, \textbf{early fusion with a lower vision ratio yields better results given a fixed total vision-text token budget}. 
This motivates our native multimodal pre-training strategy: rather than aggressive vision-heavy training concentrated at the end, we adopt a moderate vision ratio integrated early in the training process, allowing the model to naturally develop balanced multimodal representations while benefiting from extended co-optimization of both modalities.

\subsection{Zero-Vision SFT}

Pretrained VLMs do not naturally perform vision-based tool-calling, which poses a cold-start problem for multimodal RL. Conventional approaches address this issue through manually annotated or prompt-engineered chain-of-thought (CoT) data~\citep{bai2025qwen3vltechnicalreport}, but such methods are limited in diversity, often restricting visual reasoning to simple diagrams and primitive tool manipulations (\texttt{crop}, \texttt{rotate}, \texttt{flip}).

An observation is that high-quality text SFT data are relatively abundant and diverse.
We propose a novel approach, zero-vision SFT, that uses only text SFT data to activate the visual, agentic capabilities during post-training.
In this approach, all image manipulations are proxied through programmatic operations in \texttt{IPython}, effectively serving as a generalization of traditional vision tool-use. This "zero-vision" activation enables diverse reasoning behaviors, including pixel-level operations such as object size estimation via binarization and counting, and generalizes to visually grounded tasks such as object localization, counting, and OCR. 

\begin{figure}[t]
    \centering
    \includegraphics[width=0.9\textwidth]{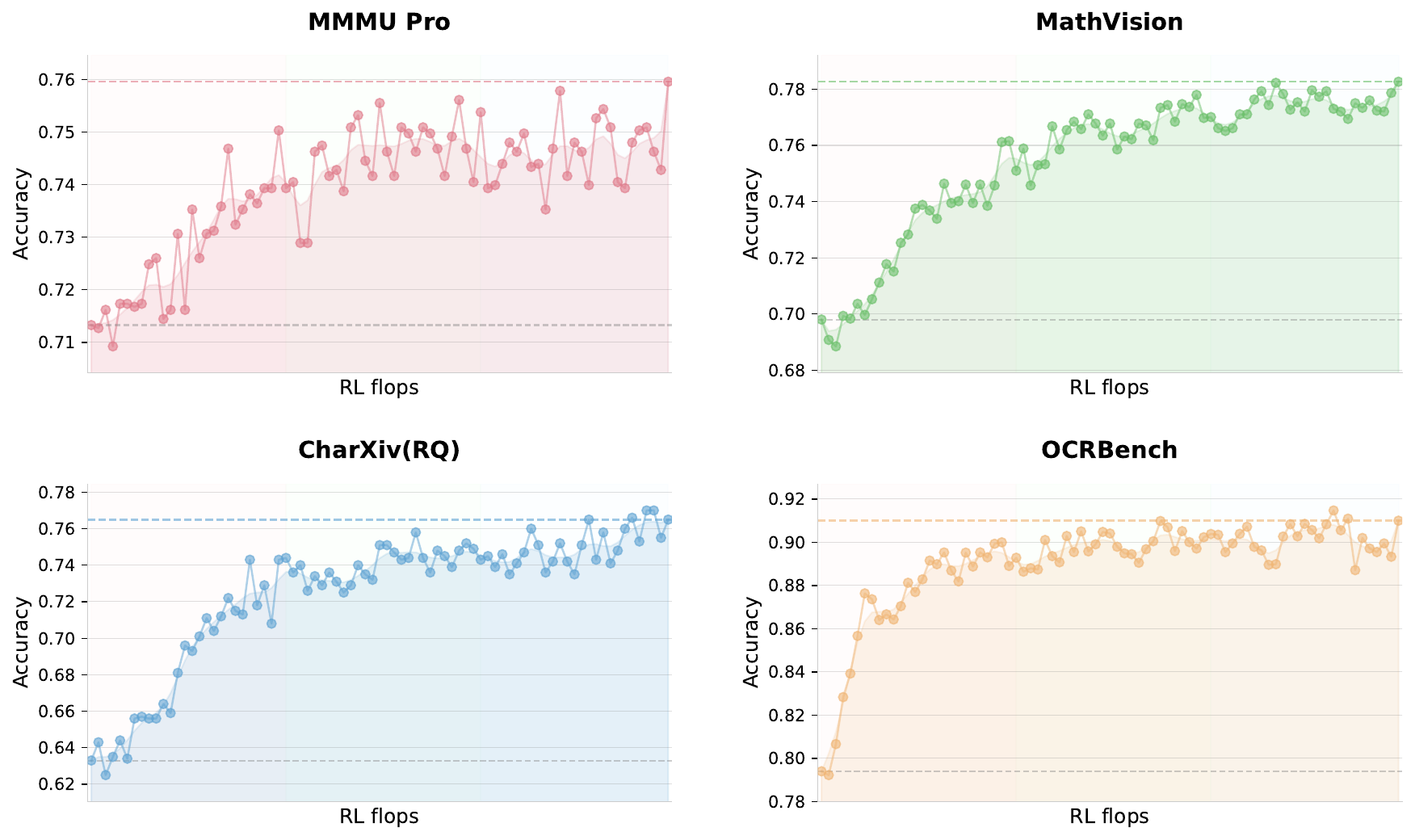}
    \caption{Vision RL training curves on vision benchmarks starting from minimal zero-vision SFT. By scaling vision RL FLOPs, the performance continues to improve, demonstrating that zero-vision activation paired with long-running RL is sufficient for acquiring robust visual capabilities.} 
    \label{fig:visionzerorl_curves}
\end{figure}

Figure~\ref{fig:visionzerorl_curves} illustrates the RL training curves, where the starting points are obtained from zero-vision SFT. The results show that zero-vision SFT is sufficient for activating vision capabilities while ensuring generalization across modalities. This phenomenon is likely due to the joint pretraining of text and vision data as described in Section~\ref{sec:joint-pre}. Compared to zero-vision SFT, our preliminary experiments show that text-vision SFT yields much worse performance on visual, agentic tasks, possibly because of the lack of high-quality vision data.

\subsection{Joint Multimodal Reinforcement Learning (RL)}
\label{sec:native_multimodal_rl}

In this section, we describe the methodology implemented in K2.5 that enables effective multimodal RL, from outcome-based visual RL to emergent cross-modal transfer that enhances textual performance.

\paragraph{Outcome-Based Visual RL}
Following the zero-vision SFT, the model requires further refinement to reliably incorporate visual inputs into reasoning. Text-initiated activation alone exhibits notable failure modes: visual inputs are sometimes ignored, and images may not be attended to when necessary. We employ outcome-based RL on tasks that explicitly require visual comprehension for correct solutions.
We categorize these tasks into three domains:
 \begin{itemize}
 \item \textbf{Visual grounding and counting:} Accurate localization and enumeration of objects within images;
 \item \textbf{Chart and document understanding:} Interpretation of structured visual information and text extraction;
 \item \textbf{Vision-critical STEM problems:} Mathematical and scientific questions filtered to require visual inputs.
 \end{itemize}
Outcome-based RL on these tasks improves both basic visual capabilities and more complex agentic behaviors. Extracting these trajectories for rejection-sampling fine-tuning (RFT) enables a self-improving data pipeline, allowing subsequent joint RL stages to leverage richer multimodal reasoning traces.

\paragraph{Visual RL Improves Text Performance}

\begin{table}[h]
\centering
\caption{Cross-Modal Transfer: Vision RL Improves Textual Knowledge}
\begin{tabular}{lccc}
\toprule
\textbf{Benchmark} & \textbf{Before Vision-RL} & \textbf{After Vision-RL} & \textbf{Improvement} \\
\midrule
MMLU-Pro & 84.7 & 86.4 & +1.7 \\
GPQA-Diamond & 84.3 & 86.4 & +2.1 \\
LongBench v2 & 56.7 & 58.9 & +2.2 \\
\bottomrule
\end{tabular}
\label{tab:vision_rl_text}
\end{table}

 To investigate potential trade-offs between visual and textual performance, we evaluated text-only benchmarks before and after visual RL. Surprisingly, outcome-based visual RL produced measurable improvements in textual tasks, including MMLU-Pro (84.7\% $\rightarrow$ 86.4\%), GPQA-Diamond (84.3\% $\rightarrow$ 86.4\%), and LongBench v2 (56.7\% $\rightarrow$ 58.9\%) (Table~\ref{tab:vision_rl_text}).
Analysis suggests that visual RL enhances calibration in areas requiring structured information extraction, reducing uncertainty on queries that resemble visually grounded reasoning (e.g., counting, OCR). These findings indicate that visual RL can contribute to cross-modal generalization, improving textual reasoning without observable degradation of language capabilities.

\textbf{Joint Multimodal RL} Motivated by the finding that robust visual capabilities can emerge from zero-vision SFT paired with vision RL---which further enhances general text abilities---we adopt a joint multimodal RL paradigm during Kimi K2.5's post-training. Departing from conventional modality-specific expert divisions, we organize RL domains not by input modality but by abilities---knowledge, reasoning, coding, agentic, etc. These domain experts jointly learn from both pure-text and multimodal queries, while the Generative Reward Model (GRM) similarly optimizes across heterogeneous traces without modality barriers. This pardaigm ensures that capability improvements acquired through either textual or visual inputs inherently generalize to enhance related abilities across the alternate modality, thereby maximizing cross-modal capability transfer.

\section{Agent Swarm}
The primary challenge of existing agent-based systems lies in their reliance on {sequential execution} of reasoning and tool-calling steps. While this structure may be effective for simpler, short-horizon tasks, it becomes inadequate as the complexity of the task increases and the accumulated context grows. As tasks evolve to contain broad information gathering and intricate, multi-branch reasoning, sequential systems often encounter significant bottlenecks \citep{anthropicbuildmultiagent,opus45report,claudemultiagent}. The limited capacity of a single agent working through each step one by one can lead to the exhaustion of practical reasoning depth and tool-call budgets, ultimately hindering the system's ability to handle more complex scenarios. 

\begin{figure}[t]
    \centering
    \includegraphics[width=\linewidth]{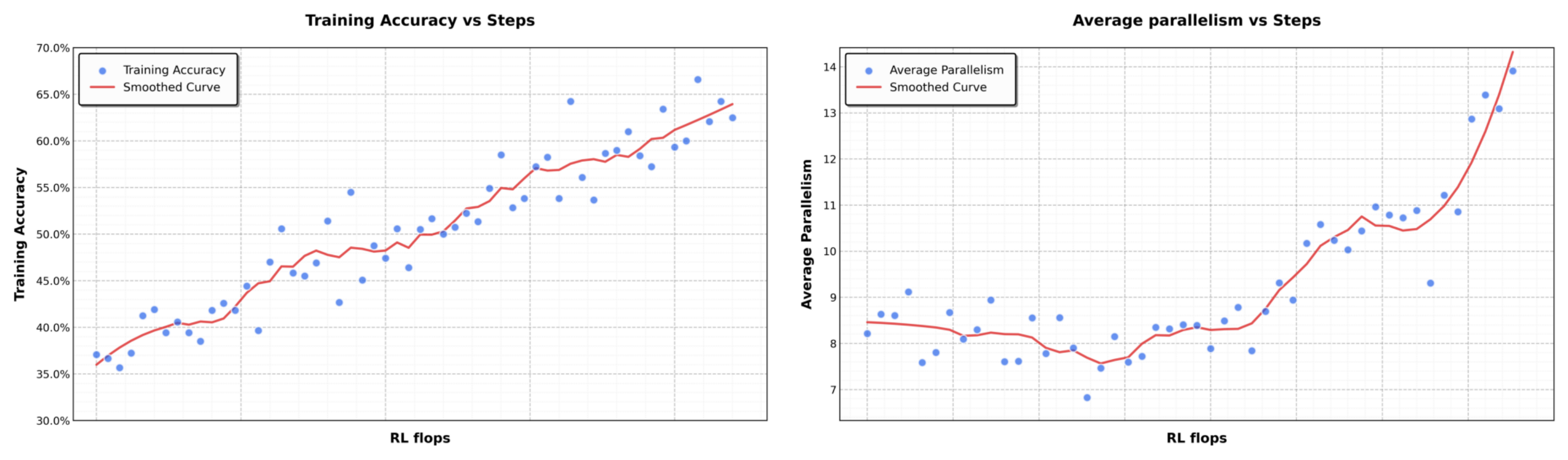}
    \caption{In our parallel-agent reinforcement learning environment, the training accuracy increases smoothly as training progresses. At the same time, the level of parallelism during training also gradually increases.}
    \label{fig:parl-learning}
\end{figure}

To address this, we introduce \textbf{Agent Swarm} and \textbf{Parallel Agent Reinforcement Learning (PARL)}. Instead of executing a task as a reasoning chain or relying on pre-specified parallelization heuristics, K2.5 initiates an Agent Swarm through dynamic task decomposition, subagent instantiation, and parallel subtask scheduling. Importantly, parallelism is not presumed to be inherently advantageous; decisions regarding whether, when, and how to parallelize are explicitly learned through environmental feedback and RL-driven exploration. 
As shown in Figure~\ref{fig:parl-learning}, the progression of performance  demonstrates this adaptive capability, with the cumulative reward increasing smoothly as the orchestrator optimizes its parallelization strategy throughout training. 

\begin{figure}[t]
    \centering
    \includegraphics[width=0.9\linewidth]{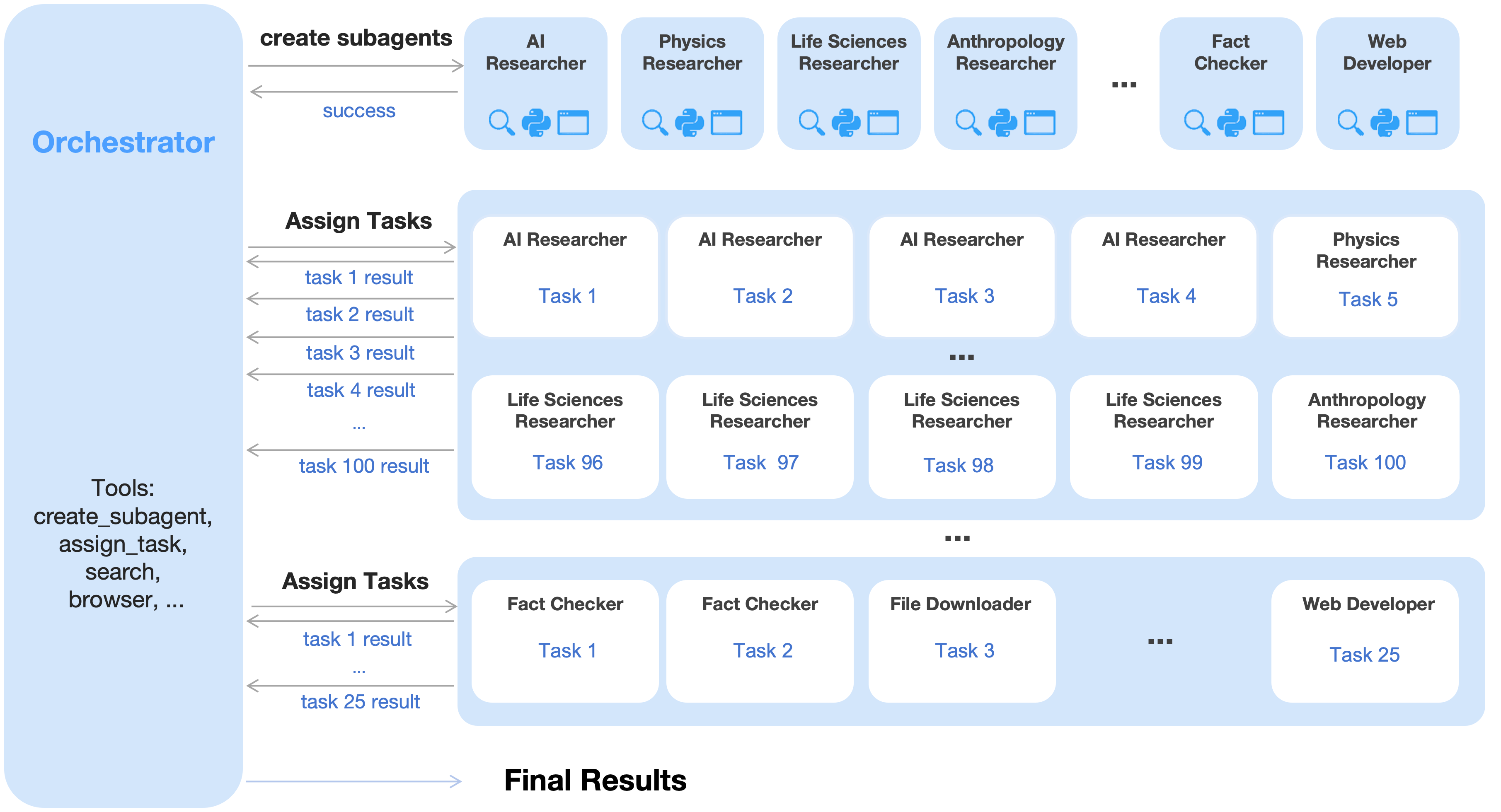}
    \caption{An agent swarm has a trainable orchestrator that dynamically creates specialized frozen subagents and decomposes complex tasks into parallelizable subtasks for efficient distributed execution.}
    \label{fig:parl-orchestration}
\end{figure}
\paragraph{Architecture and Learning Setup} 

The PARL framework adopts a decoupled architecture comprising a trainable orchestrator and frozen subagents instantiated from fixed intermediate policy checkpoints. This design deliberately avoids end-to-end co-optimization to circumvent two fundamental challenges: credit assignment ambiguity and training instability. 
In this multi-agent setting, outcome-based rewards are inherently sparse and noisy; a correct final answer does not guarantee flawless subagent execution, just as a failure does not imply universal subagent error. 
By freezing the subagents and treating their outputs as environmental observations rather than differentiable decision points, we disentangle high-level coordination logic from low-level execution proficiency, 
 leading to more robust convergence.
To improve efficiency, we first train the orchestrator using small-size subagents before transitioning to larger models.
Our RL framework also supports dynamically adjusting the inference instance ratios between subagents and the orchestrator, thereby maximizing the resource usage across the cluster.

\paragraph{PARL Reward}
Training a reliable parallel orchestrator is challenging due to the delayed, sparse, and non-stationary feedback inherent in independent subagent execution. To address this, we define the PARL reward as:
\begin{align*}
r_{\mathrm{PARL}}(x, y) = \lambda_1 \cdot \mspace{-26mu} \underbrace{r_{\text{parallel}}}_{\text{instantiation reward}} \mspace{-9mu} + \mspace{18mu}\lambda_2 \cdot \mspace{-32mu}\underbrace{r_{\text{finish}}}_{\text{sub-agent finish rate}} + 
\underbrace{r_{\text{perf}}(x, y)}_{\text{task-level outcome}}
\, .
\end{align*}

The performance reward $r_{\text{perf}}$ evaluates the overall success and quality of the solution $y$ for a given task $x$. This is augmented by two auxiliary rewards, each addressing a distinct challenge in learning parallel orchestration. The reward $r_{\text{parallel}}$ is introduced to mitigate \emph{serial collapse}—a local optimum where the orchestrator defaults to single-agent execution. By incentivizing subagent instantiation, this term encourages the exploration of concurrent scheduling spaces. The $r_{\text{finish}}$ reward focuses on the successful completion of assigned subtasks. It is used to prevent \emph{spurious parallelism}, a reward-hacking behavior in which the orchestrator increases parallel metrics dramatically by spawning many subagents without meaningful task decomposition. By rewarding completed subtasks, $r_{\text{finish}}$ enforces feasibility and guides the policy toward valid and effective decompositions.

To ensure the final policy optimizes for the primary objective, the hyperparameters $\lambda_1$ and $\lambda_2$ are annealed to zero over the course of training.

\paragraph{Critical Steps as Resource Constraint} 
To measure computational time cost in a parallel-agent setting, we define \emph{critical steps} by analogy to the \emph{critical path} in a computation graph. 
We model an episode as a sequence of execution stages indexed by $t = 1, \dots, T$. 
In each stage, the main agent executes an action, which corresponds to either direct tool invocation or the instantiation of a group of subagents running in parallel. Let $S_{\mathrm{main}}^{(t)}$ denote the number of steps taken by the main agent in stage $t$ (typically $S_{\mathrm{main}}^{(t)} = 1$), and  $S_{\mathrm{sub},i}^{(t)}$ denote the number of steps taken by the $i$-th subagent in that parallel group. 
The duration of stage $t$ is governed by the longest-running subagent within that cohort. Consequently, the total critical steps for an episode are defined as 
\begin{align*}
\text{CriticalSteps}
= \sum_{t=1}^{T} \left( S_{\mathrm{main}}^{(t)} + \max_i S_{\mathrm{sub}, i}^{(t)} \right).
\end{align*}
By constraining training and evaluation using critical steps rather than total steps, the framework explicitly incentivizes effective parallelization. Excessive subtask creation that does not reduce the maximum execution time of parallel groups yields little benefit under this metric, while well-balanced task decomposition that shortens the longest parallel branch directly reduces critical steps. As a result, the orchestrator is encouraged to allocate work across subagents in a way that minimizes end-to-end latency, rather than merely maximizing concurrency or total work performed.

\paragraph{Prompt Construction for Parallel-agent Capability Induction}
To incentivize the orchestrator to leverage the advantages of parallelization, we construct a suite of synthetic prompts designed to stress the limits of sequential agentic execution. 
These prompts emphasize either \emph{wide search}, requiring simultaneous exploration of many independent information sources, or \emph{deep search}, requiring multiple reasoning branches with delayed aggregation. We additionally include tasks inspired by real-world workloads, such as long-context document analysis and large-scale file downloading. 
When executed sequentially, these tasks are difficult to complete within fixed reasoning-step and tool-call budgets. By construction, they encourage the orchestrator to allocate subtasks in parallel, enabling completion within fewer critical steps than would be feasible for a single sequential agent. Importantly, the prompts do not explicitly instruct the model to parallelize. Instead, they shape the task distribution such that parallel decomposition and scheduling strategies are naturally favored.

\section{Method Overview}

\subsection{Foundation: Kimi K2 Base Model}

The foundation of Kimi K2.5 is Kimi K2~\citep{team2025kimik2}, a trillion-parameter mixture-of-experts (MoE) transformer~\citep{transformer} model pre-trained on 15 trillion high-quality text tokens. Kimi K2 employs the token-efficient MuonClip optimizer~\citep{jordan2024muon,liu2025muon} with QK-Clip for training stability. The model comprises 1.04 trillion total parameters with 32 billion activated parameters, utilizing 384 experts with 8 activated per token (sparsity of 48). For detailed descriptions of MuonClip, architecture design, and training infrastructure, we refer to the Kimi K2 technical report~\citep{team2025kimik2}.

\subsection{Model Architecture}

The multimodal architecture of Kimi K2.5 consists of three components: a three-dimensional native-resolution vision encoder (MoonViT-3D), an MLP projector, and the Kimi K2 MoE language model, following the design principles established in Kimi-VL~\citep{team2025kimivl}.

\paragraph{MoonViT-3D: Shared Embedding Space for Images and Videos}

In Kimi-VL, we employ MoonViT to natively process images at their original resolutions, eliminating the need for complex sub-image splitting and splicing operations. Initialized from SigLIP-SO-400M~\citep{zhai2023sigmoidlosslanguageimage}, MoonViT incorporates the patch packing strategy from NaViT~\citep{dehghani2023patchnpacknavit}, where single images are divided into patches, flattened, and sequentially concatenated into 1D sequences, thereby enabling efficient simultaneous training on images at varying resolutions.

To maximize the transfer of image understanding capabilities to video, we introduce \textbf{MoonViT-3D} with a unified architecture, fully shared parameters, and a consistent embedding space. By generalizing the ``patch n' pack`` philosophy to the temporal dimension, up to four consecutive frames are treated as a spatiotemporal volume: 2D patches from these frames are jointly flattened and packed into a single 1D sequence, allowing the identical attention mechanism to operate seamlessly across both space and time. While the extra temporal attention improves understanding on high-speed motions and visual effects, the sharing maximizes knowledge generalization from static images to dynamic videos, achieving strong video understanding performance (see in Tab.~\ref{tab:instruct_eval}) without requiring specialized video modules or architectural bifurcation. Prior to the MLP projector, lightweight temporal pooling aggregates patches within each temporal chunk, yielding $4\times$ temporal compression to significantly extend feasible video length. The result is a unified pipeline where knowledge and ability obtained from image pretraining transfers holistically to videos through one shared parameter space and feature representation.

\subsection{Pre-training Pipeline}

As illustrated in Table~\ref{tab:pretrainingdatavolume}, Kimi K2.5's pre-training builds upon the Kimi K2 language model checkpoint and processes approximately 15T tokens across three stages: first, standalone ViT training to establish a robust native-resolution visual encoder; second, joint pre-training to simultaneously enhance language and multimodal capabilities; and third, mid-training on high-quality data and long-context activation to refine capabilities and extend context windows.

\begin{table}[t]
\centering
\caption{Overview of training stages: data composition, token volumes, sequence lengths, and trainable components.}
\vspace{0.3em}
\renewcommand{\arraystretch}{1.2} 
\begin{tabular}{lccc}
\toprule
\textbf{Stages} & \textbf{ViT Training} & \textbf{Joint Pre-training} & \textbf{Joint Long-context Mid-training} \\
\midrule
\textbf{Data} & 
\makecell[c]{Alt text \\ Synthesis Caption \\ Grounding, OCR, Video} & 
\makecell[c]{+ \\ Text, Knowledge \\ Interleaving \\ Video, OS Screenshot} & 
\makecell[c]{+ \\ High-quality Text \& Multimodal \\ Long Text, Long Video \\ Reasoning, Long-CoT } \\ 
\midrule
\textbf{Sequence length} & 4096 & 4096 & 32768$\rightarrow$262144 \\
\midrule
\textbf{Tokens} & 1T & 15T & 500B$\rightarrow$200B \\
\midrule
\textbf{Training} & ViT & ViT \& LLM & ViT \& LLM \\
\bottomrule
\end{tabular}
\label{tab:pretrainingdatavolume}
\end{table}
\paragraph{ViT Training Stage}
The MoonViT-3D is continual pre-trained from SigLIP~\cite{zhai2023sigmoidlosslanguageimage} on image-text and video-text pairs, where the text components consist of a variety of targets: image alt texts, synthetic captions of images and videos, grounding bboxes, and OCR texts. Unlike the implementation in Kimi-VL~\cite{team2025kimivl}, this continual pre-training does not include a contrastive loss, but incorporates solely cross-entropy loss ${L}_{caption}$ for caption generation conditioned on input images and videos. We adopt a two-stage alignment strategy. In the first stage, we update the MoonViT-3D to align it with Moonlight-16B-A3B~\citep{liu2025muon} via the caption loss, consuming about 1T tokens with very few training FLOPs. This stage allows MoonViT-3D to primarily understand high-resolution images and videos. A very short second stage follows, updating only the MLP projector to bridge the ViT with the 1T LLM for smoother joint pre-training.

\paragraph{Joint Training Stages}
The joint pre-training stage continues from a near-end Kimi K2 checkpoint over additional 15T vision-text tokens at 4K sequence length. The data recipe extends Kimi K2's pre-training distribution by introducing unique tokens, adjusting data proportions with increased weight on coding-related content, and controlling maximum epochs per data source. The third stage performs long-context activation with integrated higher-quality mid-training data, sequentially extending context length via YaRN~\citep{peng2023yarn} interpolation. This yields significant generalization improvements in long-context text understanding and long video comprehension.

\subsection{Post-Training}
\subsubsection{Supervised Fine-Tuning}

Following the SFT pipeline established by Kimi K2 \citep{team2025kimik2}, we developed K2.5 by synthesizing high-quality candidate responses from K2, K2 Thinking and a suite of proprietary in-house expert models. Our data generation strategy employs specialized pipelines tailored to specific domains, integrating human annotation with advanced prompt engineering and multi-stage verification. 
This methodology produced a large-scale instruction-tuning dataset featuring diverse prompts and intricate reasoning trajectories, ultimately training the model to prioritize interactive reasoning and precise tool-calling for complex, real-world applications.

\subsubsection{Reinforcement Learning}
Reinforcement learning constitutes a crucial phase of our post-training. 
To facilitate joint optimization across text and vision modalities, as well as to enable PARL for agent swarm, we develop a Unified Agentic Reinforcement Learning Environment (Appendix~\ref{app:rl_infra}) and optimize the RL algorithms. 
Both text-vision joint RL and PARL are built upon the algorithms described in this section.
\paragraph{Policy Optimization}
For each problem $x$ sampled from a dataset $\mathcal{D}$, $K$ responses $\{y_1,\dots,y_K\}$ are generated using the previous policy $\pi_{\mathrm{old}}$.
We optimize the model $\pi_\theta$ with respect to the following objective: 
\begin{align}
L_{\mathrm{RL}}(\theta) = \mathbb{E}_{x \sim\mathcal{D}}
\left[ \frac{1}{N} \sum_{j=1}^K \sum_{i=1}^{|y_j|}
\mathrm{Clip}
\left(
\frac{\pi_{\theta}(y_j^i | x, y_j^{0:i})}{\pi_{\mathrm{old}}(y_j^i | x, y_j^{0:i}) }, \alpha, \beta
\right)
({r}(x, y_j) - \bar{r}(x))- \tau \left( \log \frac{\pi_{\theta}(y_j^i | x, y_j^{0:i})}{\pi_{\mathrm{old}}(y_j^i | x, y_j^{0:i}) } \right)^2
\right]
\, .
\label{eq:rl-objective}
\end{align}
Here $\alpha, \beta, \tau >0$ are hyperparameters, 
$y^j_{0:i}$ is the prefix up to the $i$-th token of the $j$-th response, 
$N=\sum_{i=1}^{K} |y_i|$ is the total number of generated tokens in a batch, 
$\bar{r}(x) = \frac{1}{K}\sum_{j=1}^K r(x, y_j)$ is the mean reward of all generated responses.   

This loss function departs from the policy optimization algorithm used in K1.5~\citep{team2025kimi} by introducing a token-level clipping mechanism designed to mitigate the off-policy divergence amplified by discrepancies between training and inference frameworks. 
The mechanism functions as a simple gradient masking scheme: policy gradients are computed normally for tokens with log-ratios within the interval $[\alpha, \beta]$, while gradients for tokens falling outside this range are zeroed out. 
Notably, a key distinction from standard PPO clipping \citep{schulman2017proximal} is that our method relies strictly on the log-ratio to explicitly bound off-policy drift, regardless of the sign of the advantages. 
This approach aligns with recent strategies proposed to stabilize large-scale RL training \citep{yao2025offpolicy, IcePop2025}. 
Empirically, we find this mechanism essential for maintaining training stability in complex domains requiring long-horizon, multi-step tool-use reasoning.
We employ the MuonClip optimizer~\citep{jordan2024muon,liu2025muon} to minimize this objective.

\paragraph{Reward Function} 
We apply a rule-based outcome reward for tasks with verifiable solutions, such as reasoning and agentic tasks. 
To optimize resource consumption, we also incorporate a budget-control reward aimed at enhancing token efficiency. 
For general-purpose tasks, we employ Generative Reward Models (GRMs) that provide granular evaluations aligned with Kimi's internal value criteria. 
In addition, for visual tasks, we design task-specific reward functions to provide fine-grained supervision. 
For visual grounding and point localization tasks, we employ an F1-based reward with soft matching: grounding tasks derive soft matches from Intersection over Union (IoU) and point tasks derive soft matches from Gaussian-weighted distances under optimal matching. 
For polygon segmentation tasks, we rasterize the predicted polygon into a binary mask and compute the segmentation IoU against the ground-truth mask to assign the reward. 
For OCR tasks, we adopt normalized edit distance to quantify character-level alignment between predictions and ground-truth. 
For counting tasks, rewards are assigned based on the absolute difference between predictions and ground-truth. 
Furthermore, we synthesize complex visual puzzle problems and utilize an LLM verifier (Kimi K2) to provide feedback.

\paragraph{Generative Reward Models}
Kimi K2 leverages a self-critique rubric reward for open-ended generation \cite{team2025kimik2}, and K2.5 extends this line of work by systematically deploying \emph{Generative Reward Models (GRMs)} across a broad range of agentic behaviors and multimodal trajectories. Rather than limiting reward modeling to conversational outputs, we apply GRMs on top of verified reward signals in diverse environments, including chat assistants, coding agents, search agents, and artifact-generating agents. Notably, GRMs function not as binary adjudicators, but as fine-grained evaluators aligned with Kimi's values that are critical to user experiences, such as helpfulness, response readiness, contextual relevance, appropriate level of detail, aesthetic quality of generated artifacts, and strict instruction following. This design allows the reward signal to capture nuanced preference gradients that are difficult to encode with purely rule-based or task-specific verifiers.
To mitigate reward hacking and overfitting to a single preference signal, we employ multiple alternative GRM rubrics tailored to different task contexts.

\paragraph{Token Efficient Reinforcement Learning}
\label{sec:te}
Token efficiency is central to LLMs with test-time scaling. 
While test-time scaling inherently trades computation for reasoning quality, practical gains require algorithmic innovations that actively navigate this trade-off. 
Our previous findings indicate that imposing a problem-dependent budget effectively constrains inference-time compute, incentivizing the model to generate more concise chain of thought reasoning patterns without unnecessary token expansion \citep{team2025kimi,team2025kimik2}. 
However, we also observe a \emph{length-overfitting phenomenon}: models trained under rigid budget constraints often fail to generalize to higher compute scales. Consequently, they cannot effectively leverage additional inference-time tokens to solve complex problems, instead defaulting to truncated reasoning patterns.

To this end, we propose \emph{Toggle}, 
a training heuristic that alternates between {inference-time scaling} and {budget-constrained optimization}: 
for learning iteration $t$, 
the reward function is defined by 
\begin{align*}
\tilde{r}(x,y) = 
\begin{cases} 
r(x, y) \cdot \mathbb{I}\left\{ \frac{1}{K} \sum_{i=1}^K r(x, y_i)  < \lambda\ \mathrm{or}\ |y_i| \leq \mathrm{budget(x)} \right\} & \text{if } \lfloor t/m \rfloor \pmod 2 = 0\ (\mathrm{{Phase 0}}) \\
r(x, y) & \text{if } \lfloor t/m \rfloor \pmod 2 = 1\ (\mathrm{{Phase 1}}) 
\end{cases}
\, .
\end{align*}
where $\lambda$ and $m$ are hyper-parameters of the algorithm and $K$ is the number of rollouts per problem. 
Specifically, the algorithm alternates between two optimization phases every $m$ iterations:
\begin{itemize}

\item {Phase0} (\emph{budget limited phase}): 
The model is trained to solve the problem within a task-dependent token budget. 
To prevent a premature sacrifice of quality for efficiency, this constraint is conditionally applied: it is only enforced when the model's mean accuracy for a given problem exceeds the threshold $\lambda$.

\item {Phase1} (\emph{standard scaling phase}): The model generates responses up to the maximum token limit, encouraging the model to leverage computation for better inference-time scaling. 

\end{itemize}

The problem-dependent budget is estimated from the $\rho$-th percentile of token lengths among the subset of correct responses:
\begin{equation}
\mathrm{budget}(x) = \text{Percentile}\left(\{ |y_j| \mid r(x, y_i) = 1, i=1,\dots, K \}, \rho \right) \,.
\end{equation}
This budget is estimated once at the beginning of training and remains fixed thereafter. 
Notably, Toggle functions as a stochastic alternating optimization for a bi-objective problem. It is specifically designed to reconcile reasoning capabilities with computational efficiency. 

\begin{figure}[t]
    \centering
\includegraphics[width=0.8\linewidth, keepaspectratio]{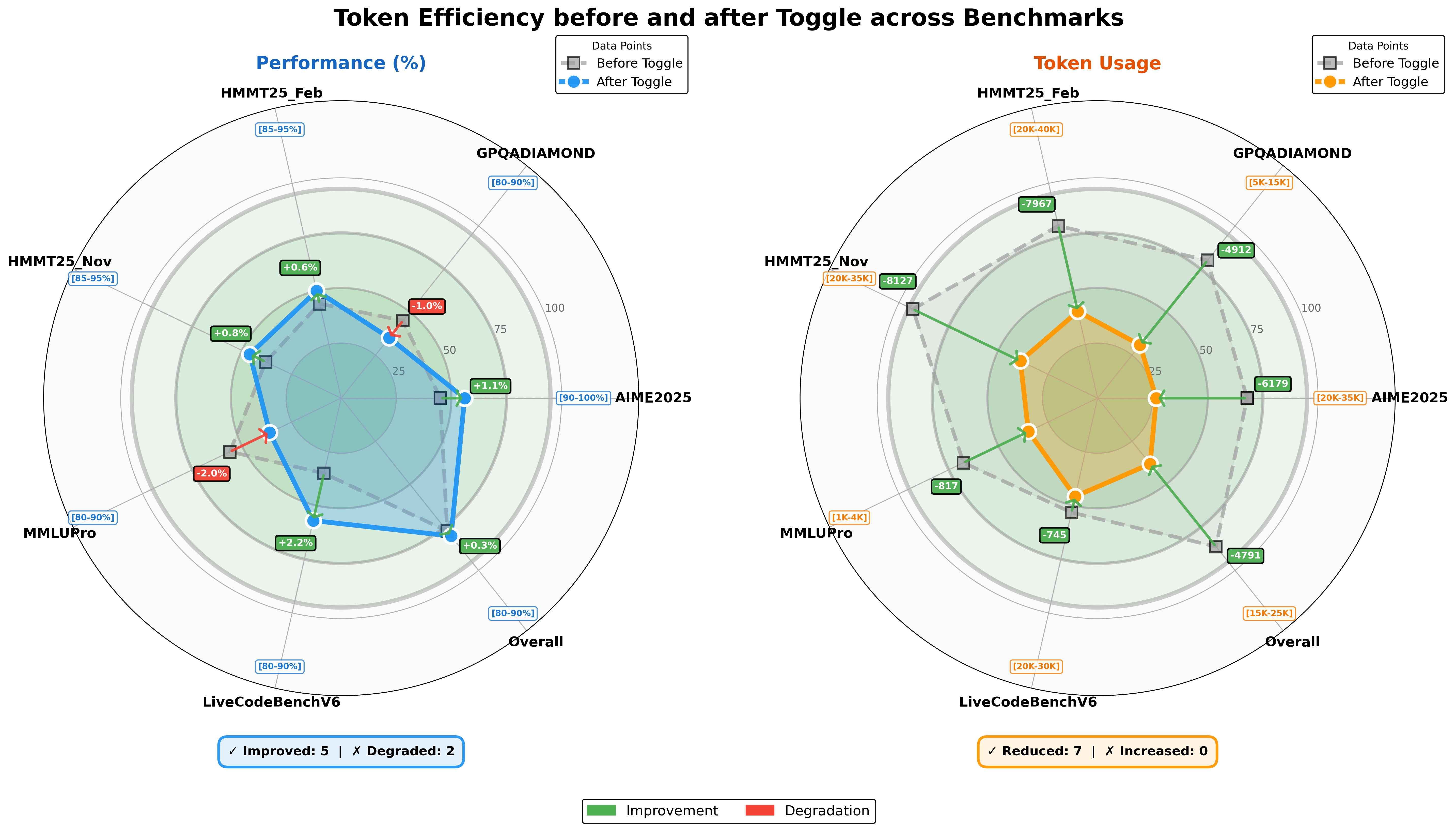}
    \caption{Comparison of model performance and token usage for Kimi K2 Thinking following token-efficient RL.}
    \label{fig:k25_performance}
\end{figure}

We evaluate the effectiveness of Toggle on K2 Thinking~\citep{moonshotai2025kimi}.
As shown in Figure~\ref{fig:k25_performance}, we observe a consistent reduction in output length across nearly all benchmarks.
On average, Toggle decreases output tokens by 25$\sim$30\% with a negligible impact on performance. 
We also observe that redundant patterns in the chain-of-thought, such as repeated verifications and mechanical calculations, decrease substantially.
Furthermore, Toggle shows strong domain generalization. 
For example, when trained exclusively on mathematics and programming tasks, the model still achieves consistent token reductions on GPQA and MMLU-Pro with only marginal degradation in performance (Figure \ref{fig:k25_performance}).

\subsection{Training Infrastructure}

Kimi K2.5 inherits the training infrastructure from Kimi K2~\citep{team2025kimik2} with minimal modifications.  For multimodal training, we propose Decoupled Encoder Process, where the vision encoder is incorporated into the existing pipeline with negligible additional overhead.

\subsubsection{Decoupled Encoder Process (DEP)}

In a typical multimodal training paradigm utilizing Pipeline Parallelism (PP), the vision encoder and text embedding are co-located in the first stage of the pipeline (Stage-0). However, due to the inherent variations of multimodal input size (e.g., image counts and resolutions), Stage-0 suffers from drastic fluctuations in both computational load and memory usage. This forces existing solutions to adopt custom PP configurations for vision-language models --- for instance, \cite{team2025kimivl} manually adjusts the number of text decoder layers in Stage-0 to reserve memory. While this compromise alleviates memory pressure, it does not fundamentally resolve the load imbalance caused by multimodal input sizes. More critically, it precludes the direct reuse of parallel strategies that have been highly optimized for text-only training.

Leveraging the unique topological position of the visual encoder within the computation graph --- specifically, its role as the start of the forward pass and the end of the backward pass --- our training uses \textbf{Decoupled Encoder Process (DEP)}, which is composed of three stages in each training step:

\begin{itemize}
\item \textbf{Balanced Vision Forward:} We first execute the forward pass for all visual data in the global batch. Because the vision encoder is small, we replicate it on all GPUs regardless of other parallelism strategies. During this phase, the forward computational workload is evenly distributed across all GPUs based on load metrics (e.g., image or patch counts). This eliminates load-imbalance caused by PP and visual token counts. To minimize peak memory usage, we discard all intermediate activations, retaining only the final output activations. The results are gathered back to PP Stage-0;

\item \textbf{Backbone Training:} This phase performs the forward and backward passes for the main transformer backbone. By discarding intermediate activations in the preceding phase, we can now fully leverage any efficient parallel strategies validated in pure text training. After this phase, gradients are accumulated at the visual encoder output;
\item \textbf{Vision Recomputation \& Backward:} We re-compute the vision encoder forward pass, followed by a backward pass to compute gradients for parameters in the vision encoder;
\end{itemize}

DEP not only achieves load-balance, but also decouples the optimization strategy of the vision encoder and the main backbone. K2.5 seamlessly inherits the parallel strategy of K2, achieving a multimodal training efficiency of 90\% relative to text-only training. We note a concurrent work, LongCat-Flash-Omni~\citep{team2025longcat}, shares a similar design philosophy.

\section{Evaluations}
\subsection{Main Results}
\subsubsection{Evaluation Settings}
\paragraph{Benchmarks}
We evaluate Kimi K2.5 on a comprehensive benchmark suite spanning text-based reasoning, competitive and agentic coding, multimodal understanding (image and video), autonomous agentic execution, and computer use. Our benchmark taxonomy is organized along the following capability axes:
\begin{itemize}
\item \textbf{Reasoning \& General}: Humanity's Last Exam (HLE)~\citep{phan2025humanitysexam}, AIME 2025~\citep{aime2025}, HMMT 2025 (Feb)~\citep{hmmt2025feb}, IMO-AnswerBench~\citep{luong-etal-2025-towards}, GPQA-Diamond~\citep{rein2024gpqa}, MMLU-Pro~\citep{wang2024mmluprorobustchallengingmultitask}, SimpleQA Verified~\citep{haas2025simpleqaverifiedreliablefactuality}, AdvancedIF~\citep{he2025advancedifrubricbasedbenchmarkingreinforcement}, and LongBench v2~\citep{bai2025longbenchv2deeperunderstanding}.
\item \textbf{Coding}: 
SWE-Bench Verified~\citep{jimenez2023swe}, 
SWE-Bench Pro (public)~\citep{deng2025swe}, 
SWE-Bench Multilingual~\citep{jimenez2023swe}, 
Terminal Bench~2.0~\citep{merrill2026terminal}, 
PaperBench (CodeDev)~\citep{starace2025paperbench}, 
CyberGym~\citep{wang2025cybergym}, 
SciCode~\citep{tian2024scicode}, 
OJBench (cpp)~\citep{wang2025ojbench}, 
and LiveCodeBench (v6)~\citep{jain2024livecodebench}.
\item \textbf{Agentic Capabilities}: BrowseComp~\cite{wei2025browsecomp}, WideSearch~\cite{wong2025widesearch},DeepSearchQA~\cite{vedula2025deepsearchqa}, FinSearchComp (T2\&T3)~\cite{hu2025finsearchcomp}, Seal-0~\cite{pham2025sealqa}, GDPVal~\cite{patwardhan2025gdpval}.
\item \textbf{Image Understanding}: \textit{(math \& reasoning)} MMMU-Pro~\citep{yue2025mmmupro}, MMMU (val)~\citep{yue2023mmmu}, CharXiv (RQ)~\citep{wang2024charxiv}, MathVision~\citep{wang2024mathvision} and MathVista (mini)~\citep{lu2024mathvista}; \textit{(vision knowledge)} SimpleVQA~\citep{cheng2025simplevqa} and WorldVQA~\footnote{https://github.com/MoonshotAI/WorldVQA}; \textit{(perception)} ZeroBench (w/ and w/o tools)~\citep{roberts2025zerobench}, BabyVision~\citep{chen2026babyvision}, BLINK~\citep{fu2024blinkmultimodallargelanguage} and MMVP~\citep{tong2024eyeswideshutexploring}; \textit{(OCR \& document)} OCRBench~\citep{liu2024ocrbench}, OmniDocBench~1.5~\citep{ouyang2025omnidocbench} and InfoVQA~\citep{mathew2021infovqa}.
\item \textbf{Video Understanding}: VideoMMMU~\citep{hu2025videommmu}, MMVU~\citep{zhao2025mmvu}, MotionBench~\citep{hong2025motionbench}, Video-MME~\citep{fu2025videomme} (\textit{with subtitles}), LongVideoBench~\citep{wu2024longvideobench}, and LVBench~\citep{wang2025lvbench}.
\item \textbf{Computer Use}:  OSWorld-Verified~\cite{osworld_verified, OSWorld}, and WebArena ~\citep{zhou2023webarena}.
\end{itemize}

\newcolumntype{P}[1]{>{\centering\arraybackslash}p{#1}}
\begin{table}[htbp]
\vspace{-2em} 
\centering
\footnotesize
\setlength{\tabcolsep}{3pt}

\caption{Performance comparison of Kimi K2.5 against open-source and proprietary models. \textbf{Bold} denotes the global SOTA; Data points marked with * are taken from our internal evaluations. $^{\dagger}$ refers to their scores of text-only subset.}
\label{tab:instruct_eval}

\begin{tabular}{@{}l | P{1.5cm} | P{2cm} P{2cm} P{2cm} | P{2cm} P{2cm}@{}} 
\toprule
& & \multicolumn{3}{c|}{\textbf{Proprietary}} & \multicolumn{2}{c}{\textbf{Open Source}} \\ 
\cmidrule(l{2pt}r{2pt}){3-5} \cmidrule(l{2pt}r{2pt}){6-7}
\textbf{Benchmark} & \textbf{Kimi K2.5} & \textbf{Claude Opus 4.5} & \textbf{GPT-5.2 (xhigh)} & \textbf{Gemini 3 Pro} & \textbf{DeepSeek-V3.2} & \textbf{Qwen3-VL-235B-A22B} \\
\midrule
\multicolumn{7}{@{}l}{\textbf{Reasoning \& General}} \\
HLE-Full & 30.1 & 30.8 & 34.5 & \textbf{37.5} & 25.1$^{\dagger}$ & - \\
HLE-Full~w/ tools & \textbf{50.2 }& 43.2 & 45.5 & 45.8 & 40.8$^{\dagger}$ & - \\
AIME~2025 & 96.1 & 92.8 & \textbf{100} & 95.0 & 93.1 & - \\
HMMT~2025 (Feb) & 95.4 & 92.9* & \textbf{99.4} & 97.3* & 92.5 & - \\
IMO-AnswerBench & 81.8 & 78.5* & \textbf{86.3} & 83.1* & 78.3 & - \\
GPQA-Diamond & 87.6 & 87.0 & \textbf{92.4} & 91.9 & 82.4 & - \\
MMLU-Pro & 87.1 & 89.3* & 86.7* & \textbf{90.1} & 85.0 & - \\
SimpleQA Verified & 36.9 & 44.1 & 38.9 & \textbf{72.1} & 27.5 & - \\
AdvancedIF & 75.6 & 63.1 & \textbf{81.1} & 74.7 & 58.8 & - \\
LongBench v2 & 61.0 & 64.4* & 54.5* & \textbf{68.2*} & 59.8* & - \\
\midrule

\multicolumn{7}{@{}l}{\textbf{Coding}} \\
SWE-Bench Verified & 76.8 & \textbf{80.9} & 80.0 & 76.2 & 73.1 & - \\
SWE-Bench Pro (public) & 50.7 & 55.4* & \textbf{55.6} & - & - & - \\
SWE-Bench Multilingual & 73.0 & \textbf{77.5} & 72.0 & 65.0 & 70.2 & - \\
Terminal Bench~2.0 & 50.8 & \textbf{59.3} & 54.0 & 54.2 & 46.4 & - \\
PaperBench (CodeDev) & 63.5 & \textbf{72.9*} & 63.7* & - & 47.1 & - \\
CyberGym & 41.3 & \textbf{50.6} & - & 39.9* & 17.3* & - \\
SciCode & 48.7 & 49.5 & 52.1 & \textbf{56.1} & 38.9 & - \\
OJBench (cpp) & 57.4 & 54.6* & - & \textbf{68.5*} & 54.7* & - \\
LiveCodeBench (v6) & 85.0 & 82.2* & - & \textbf{87.4*} & 83.3 & - \\
\midrule
\multicolumn{7}{@{}l}{\textbf{Agentic}} \\
BrowseComp & \textbf{60.6} & 37.0 & \multirow{2}{*}{65.8} & 37.8 & 51.4 & - \\
BrowseComp~(w/ ctx manage) & \textbf{74.9} & 57.8 & & 59.2 & 67.6 & - \\
BrowseComp~(Agent Swarm) & \textbf{78.4} & - & - & - & - & - \\
WideSearch & 72.7 & \textbf{76.2*} & - & 57.0 & 32.5* & - \\
WideSearch~(Agent Swarm) & \textbf{79.0} & - & - & - & - & - \\
DeepSearchQA & \textbf{77.1} & 76.1* & 71.3* & 63.2* & 60.9* & - \\
FinSearchCompT2\&T3 & \textbf{67.8} & 66.2* & - & 49.9 & 59.1* & - \\
Seal-0 & \textbf{57.4} & 47.7* & 45.0 & 45.5* & 49.5* & - \\
GDPVal-AA & 41.0 & 45.0 & \textbf{48.0} & 35.0 & 34.0 & - \\

\midrule
\multicolumn{7}{@{}l}{\textbf{Image}} \\
MMMU-Pro & 78.5 & 74.0 & 79.5* & \textbf{81.0} & - & 69.3 \\
MMMU (val) & 84.3 & 80.7 & 86.7* & \textbf{87.5*} & - & 80.6 \\
CharXiv (RQ) & 77.5 & 67.2* & \textbf{82.1} & 81.4 & - & 66.1 \\
MathVision & 84.2 & 77.1* & 83.0 & \textbf{86.1*} & - & 74.6 \\
MathVista (mini) & \textbf{90.1} & 80.2* & 82.8* & 89.8* & - & 85.8 \\
SimpleVQA & \textbf{71.2} & 69.7* & 55.8* & 69.7* & - & 56.8* \\
WorldVQA & 46.3 & 36.8 & 28.0 & \textbf{47.4} & - & 23.5 \\
ZeroBench & \textbf{9} & 3* & \textbf{9*} & 8* & - & 4* \\
ZeroBench~w/ tools & 11 & 9* & 7* & \textbf{12*} & - & 3* \\
BabyVision & 36.5 & 14.2 & 34.4 & \textbf{49.7} & - & 22.2 \\
BLINK & \textbf{78.9}  & 68.8* & -  & 78.7* & - & 68.9 \\
MMVP & 87.0 & 80.0* & 83.0* & \textbf{90.0*} & - & 84.3 \\
OmniDocBench~1.5 & \textbf{88.8} & 87.7* & 85.7 & 88.5 & - & 82.0* \\
OCRBench & \textbf{92.3} & 86.5* & 80.7* & 90.3* & - & 87.5 \\
InfoVQA (test) & \textbf{92.6} & 76.9* & 84* & 57.2* & - & 89.5 \\

\midrule
\multicolumn{7}{@{}l}{\textbf{Video}} \\
VideoMMMU & 86.6 & 84.4* & 85.9 & \textbf{87.6} & - & 80.0 \\
MMVU & 80.4 & 77.3* & \textbf{80.8*} & 77.5* & - & 71.1 \\
MotionBench & \textbf{70.4} & 60.3* & 64.8* & 70.3 & - & - \\
Video-MME & 87.4 & 77.6* & 86.0* & \textbf{88.4*} & - & 79.0 \\
LongVideoBench & \textbf{79.8} & 67.2* & 76.5* & 77.7* & - & 65.6* \\
LVBench & \textbf{75.9} & 57.3 & - & 73.5* & - & 63.6 \\
\midrule

\multicolumn{7}{@{}l}{\textbf{Computer Use}} \\

OSWorld-Verified &  63.3 &  \textbf{66.3}& 8.6* & 20.7* & - & 38.1 \\
WebArena &  58.9 & \textbf{63.4}* & - & - & - & 26.4* \\

\bottomrule
\end{tabular}
\end{table}

\begin{table}[htbp]
\centering
\hspace{0.3cm}
\caption{Performance and token efficiency of some reasoning models. Average output token counts (in thousands) are shown in parentheses.}
\begin{tabular}{cc|ccc}
\toprule
\textbf{Benchmark} & \textbf{Kimi K2.5} & \textbf{Kimi K2} & \textbf{Gemini-3.0} & \textbf{DeepSeek-V3.2} \\
 &  & \textbf{Thinking} & \textbf{Pro} & \textbf{Thinking} \\
\midrule
AIME 2025 & 96.1 (25k) & 94.5 (30k) & 95.0 (15k) & 93.1 (16k) \\
HMMT Feb 2025 & 95.4 (27k) & 89.4 (35k) & 97.3 (16k) & 92.5 (19k) \\
HMMT Nov 2025 & 91.1 (24k) & 89.2 (32k) & 94.5 (15k) & 90.2 (18k) \\
IMO-AnswerBench & 81.8 (36k) & 78.6 (37k) & 83.1 (18k) & 78.3 (27k) \\
LiveCodeBench & 85.0 (18k) & 82.6 (25k) & 87.4 (13k) & 83.3 (16k) \\
GPQA Diamond & 87.6 (14k) & 84.5 (13k) & 91.9 (8k) & 82.4 (7k) \\
HLE-Text  & 31.5 (24k) & 23.9 (29k) & 38.4 (13k) & 25.1 (21k) \\
\bottomrule
\end{tabular}
\label{tab:k2.5-te}
\end{table}

\paragraph{Baselines}
We benchmark against state-of-the-art proprietary and open-source models. For proprietary models, we compare against Claude Opus 4.5 (with extended thinking) \citep{opus45report}, GPT-5.2 (with xhigh reasoning effort) \citep{gpt52report}, and Gemini 3 Pro (with high reasoning-level) \citep{gemini3pro}. For open-source models, we include DeepSeek-V3.2 (with thinking mode enabled)~\citep{deepseekai2025deepseekv32pushingfrontieropen} for text benchmarks, while vision benchmarks report Qwen3-VL-235B-A22B-Thinking~\citep{bai2025qwen3vltechnicalreport} instead.

\paragraph{Evaluation Configurations}
Unless otherwise specified, all Kimi K2.5 evaluations use temperature = 1.0, top-p = 0.95, and a context length of 256k tokens. Benchmarks without publicly available scores were re-evaluated under identical conditions and marked with an asterisk (*). The full evaluation settings can be found in appendix~\ref{app:eval_details}.

\subsubsection{Evaluation Results}
\label{sec:eval_results}
Comprehensive results comparing Kimi K2.5 against proprietary and open-source baselines are presented in Table~\ref{tab:instruct_eval}. We highlight key observations across core capability domains:
\paragraph{Reasoning and General}
Kimi K2.5 achieves competitive performance with top-tier proprietary models on rigorous STEM benchmarks. On Math tasks, AIME 2025, K2.5 scores 96.1\%, approaching GPT-5.2's perfect score while outperforming Claude Opus 4.5 (92.8\%) and Gemini 3 Pro (95.0\%). This high-level performance extends to the HMMT 2025 (95.4\%) and IMO-AnswerBench (81.8\%), demonstrating K2.5's superior reasoning depth. Kimi K2.5 also exhibits remarkable knowledge and scientific reasoning capabilities, scoring 36.9\% on SimpleQA Verified, 87.1\% on MMLU-Pro and 87.6\% on GPQA. Notably, on HLE without the use of tools, K2.5 achieves an HLE-Full score of 30.1\%, with component-wise scores of 31.5\% on text subset and 21.3\% on image subset. When tool-use is enabled, K2.5’s HLE-Full score rises to 50.2\%, with 51.8\% (text) and 39.8\% (image), significantly outperforming Gemini 3 Pro (45.8\%) and GPT-5.2 (45.5\%). In addition to reasoning and knowledge, K2.5 shows strong instruction-following performance (75.6\% on AdvancedIF) and competitive long-context abilities, achieving 61.0\% on LongBench v2 compared to both proprietary and open-source models.

\paragraph{Complex Coding and Software Engineering}
Kimi K2.5 exhibits strong software engineering capabilities, especially on realistic coding and maintenance tasks. It achieves 76.8\% on SWE-Bench Verified and 73.0\% on SWE-Bench Multilingual, outperforming Gemini 3 Pro while remaining competitive with Claude Opus 4.5 and GPT‑5.2. On LiveCodeBench v6, Kimi K2.5 reaches 85.0\%, surpassing DeepSeek‑V3.2 (83.3\%) and Claude Opus 4.5 (82.2\%), highlighting its robustness on live, continuously updated coding challenges. On TerminalBench 2.0, PaperBench, and SciCode, it scores 50.8\%, 63.5\%, and 48.7\% respectively, demonstrating stable competition‑level performance in automated software engineering and problem solving across diverse domains. In addition, K2.5 attains a score of 41.3 on CyberGym, on the task of finding previously discovered vulnerabilities in real open‑source software projects given only a high‑level description of the weakness, further underscoring its effectiveness in security‑oriented software analysis.

\paragraph{Agentic Capabilities}
Kimi K2.5 establishes new state-of-the-art performance on complex agentic search and browsing tasks. On BrowseComp, K2.5 achieves 60.6\% without context management techniques, 74.9\% with Discard-all context management \citep{deepseekai2025deepseekv32pushingfrontieropen} — substantially outperforming GPT-5.2's reported 65.8\%, Claude Opus 4.5 (37.0\%) and Gemini 3 Pro (37.8\%). Similarly, WideSearch reaches 72.7\% on item-f1. On DeepSearchQA (77.1\%), FinSearchCompT2\&T3 (67.8\%) and Seal-0 (57.4\%), K2.5 leads all evaluated models, demonstrating superior capacity for agentic deep research, information synthesis, and multi-step tool orchestration.

\paragraph{Vision Reasoning, Knowledge and Perception} Kimi K2.5 demonstrates strong visual reasoning and world knowledge capabilities. It scores 78.5\% on MMMU-Pro, spanning multi-disciplinary multimodal tasks. For world knowledge question answering, K2.5 achieves 71.2\% on SimpleVQA and 46.3\% on WorldVQA. For visual reasoning, it achieves 84.2\% on MathVision, 90.1\% on MathVista (mini), and 36.5\% on BabyVision. For OCR and document understanding, K2.5 delivers outstanding results with 77.5\% on CharXiv (RQ), 92.3\% on OCRBench, 88.8\% on OmniDocBench 1.5, and 92.6\% on InfoVQA (test). On the challenging ZeroBench, Kimi K2.5 achieves 9\% and 11\% with tool augmentation, substantially ahead of competing models. On basic visual perception benchmarks BLINK (78.9\%) and MMVP (87.0\%), we also observe competitive performance of Kimi K2.5, demonstrating its robust real-world visual perceptions.

\paragraph{Video Understanding} Kimi K2.5 achieves state-of-the-art performance across diverse video understanding tasks. It attains 86.6\% on VideoMMMU and 80.4\% on MMVU, rivaling frontier leaderships. With the context-compression and dense temporal understanding abilities of MoonViT-3D, Kimi K2.5 also establishes new global SOTA records in long-video comprehension with 75.9\% on LVBench and 79.8\% on LongVideoBench by feeding over 2,000 frames, while demonstrating robust dense-motion understanding at 70.4\% on the highly-dimensional MotionBench.

\paragraph{Computer-Use Capability} Kimi K2.5 demonstrates state-of-the-art computer-use capability on real-world tasks. On the computer-use benchmark OSWorld-Verified~\citep{osworld_verified,OSWorld}, it achieves a 63.3\% success rate relying solely on GUI actions without external tools. This substantially outperforms open-source models such as Qwen3-VL-235B-A22B (38.1\%) and OpenAI’s computer-use agent framework Operator (o3-based) (42.9\%), while remaining competitive with the current leading CUA model, Claude Opus 4.5 (66.3\%). On WebArena~\citep{zhou2023webarena}, an established benchmark for GUI-based web browsing, Kimi K2.5 achieves a 58.9\% success rate, surpassing OpenAI's Operator (58.1\%) and approaching the performance of Claude Opus 4.5 (63.4\%).
\subsection{Agent Swarm Results}
\label{sec:agent_swarm_results}

\paragraph{Benchmarks}
To rigorously evaluate the effectiveness of the agent swarm framework, we select three representative benchmarks that collectively cover deep reasoning, large-scale retrieval, and real-world complexity:

\begin{itemize}
\item \textbf{BrowseComp}: A challenging deep-research benchmark that requires multi-step reasoning and complex information synthesis.

\item \textbf{WideSearch}: A benchmark designed to evaluate the ability to perform broad, multi-step information seeking and reasoning across diverse sources.

\item \textbf{In-house Swarm Bench}: An internally developed Swarm benchmark, designed to evaluate the agent swarm performance under real-world, high-complexity conditions. It covers four domains: \textit{WildSearch} (unconstrained, real-world information retrieval over the open web), \textit{Batch Download} (large-scale acquisition of diverse resources), \textit{WideRead} (large-scale document comprehension involving more than 100 input documents), and \textit{Long-Form Writing} (coherent generation of extensive content exceeding 100k words). This benchmark incorporates extreme-scale scenarios that stress-test the orchestration, scalability, and coordination capabilities of agent-based systems.
\end{itemize}

\begin{table}[t]
\centering
\footnotesize
\setlength{\tabcolsep}{6pt}
\hspace{0.5cm}
\caption{Performance comparison of Kimi K2.5 Agent Swarm against single-agent and proprietary baselines on agentic search benchmarks. \textbf{Bold} denotes the best result per benchmark.}
\label{tab:agent_swarm_results}
\begin{tabular}{@{}wc{4cm} wc{2cm} wc{2cm}wc{2cm}wc{2cm}wc{2cm}@{}} 
\toprule
\textbf{Benchmark} & \textbf{K2.5 Agent Swarm} & \textbf{Kimi K2.5} & \textbf{Claude Opus 4.5} & \textbf{GPT-5.2} & \textbf{GPT-5.2 Pro} \\
\midrule
BrowseComp & \textbf{78.4} & 60.6 & 37.0 & 65.8 & 77.9 \\
WideSearch & \textbf{79.0} & 72.7 & 76.2 & - & - \\
In-house Swarm Bench & \textbf{58.3} & 41.6 & 45.8 & - & - \\
\bottomrule
\end{tabular}
\end{table}
\paragraph{Performance} Table~\ref{tab:agent_swarm_results} presents the performance of Kimi K2.5 Agent Swarm against single-agent configurations and proprietary baselines. The results demonstrate substantial performance improvements from multi-agent orchestration. On BrowseComp, Agent Swarm achieves 78.4\%, representing a 17.8\% absolute gain over the single-agent K2.5 (60.6\%) and surpassing even GPT-5.2 Pro (77.9\%). Similarly, WideSearch sees a 6.3\% improvement (72.7\% $\to$ 79.0\%) on Item-F1, enabling K2.5 Agent Swarm to outperform Claude Opus 4.5 (76.2\%) and establish a new state-of-the-art. The gains are most pronounced on In-house Swarm bench (16.7\%), where tasks are explicitly designed to reward parallel decomposition. These consistent improvements across benchmarks validate that Agent Swarm effectively converts computational parallelism into qualitative capability gains, particularly for problems requiring broad exploration, multi-source verification, or simultaneous handling of independent sub-tasks.
\begin{figure}[t]
    \centering
    \begin{minipage}[t]{0.48\linewidth}
        \centering
        \includegraphics[width=0.95\linewidth, keepaspectratio]{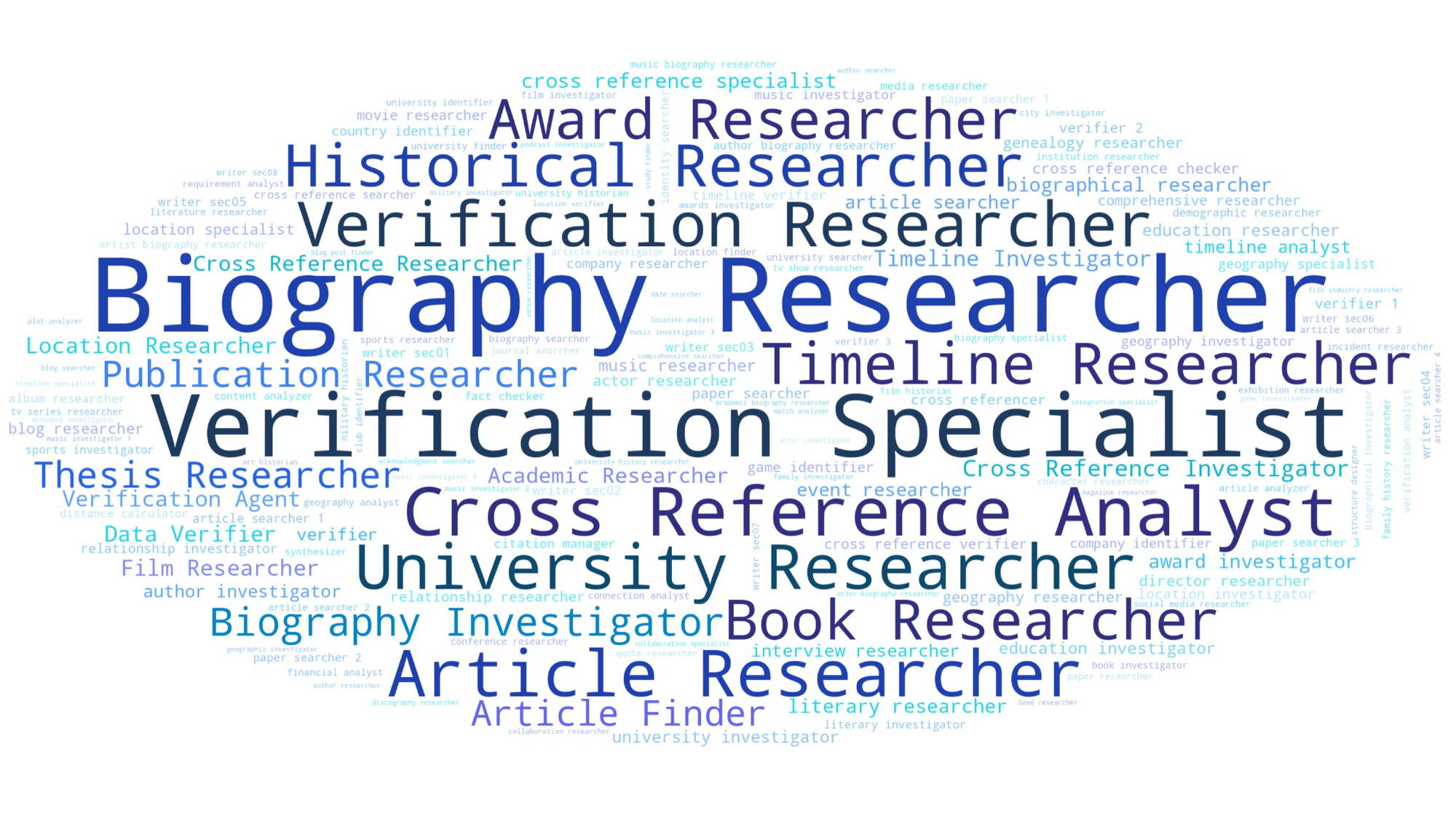}
        \caption{The word cloud visualizes heterogeneous K2.5-based sub-agents dynamically instantiated by the Orchestrator across tests.}
        \label{fig:subagent_in_agent_swarm}
    \end{minipage}
    \hfill
    \begin{minipage}[t]{0.48\linewidth}
        \centering
        \includegraphics[width=0.95\linewidth, keepaspectratio]{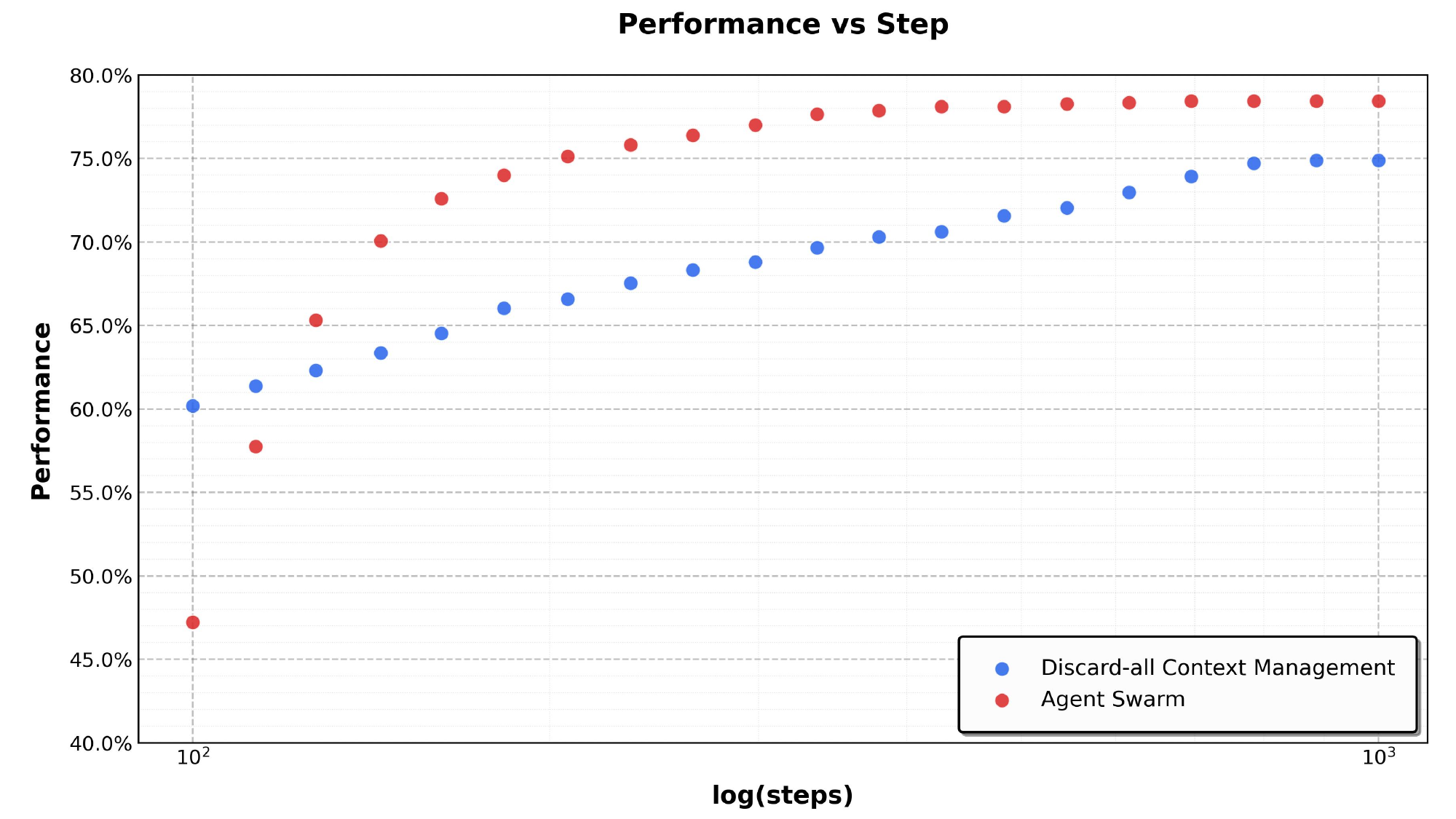}
        \caption{Comparison of Kimi K2.5 performance under Agent Swarm and Discard-all context management in BrowseComp.}
        \label{fig:agent_swarm_as_ctx_mgm}
    \end{minipage}
\end{figure}

\begin{figure}[t]
    \centering
\includegraphics[width=0.618\linewidth, keepaspectratio]{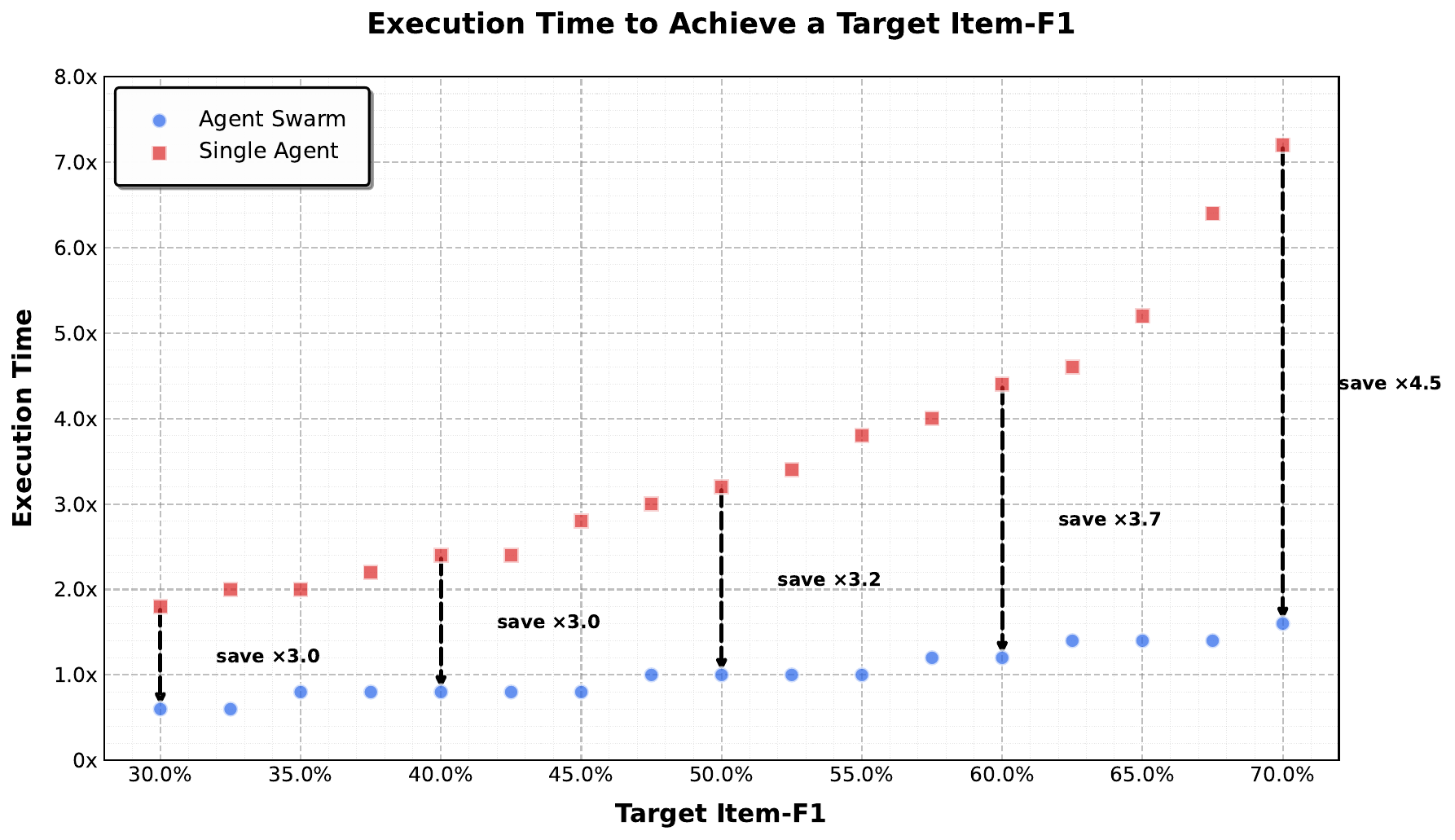}
    \caption{Agent Swarm achieves 3$\times$--4.5$\times$ faster execution time compared to single-agent baselines as target Item-F1 increases from 30\% to 70\% in WideSearch testing.}
    \label{fig:agent_swarm_efficiency}
\end{figure}

\paragraph{Execution Time Savings via Parallelism} Beyond improved task performance, Agent Swarm achieves substantial wall-clock time reductions through parallel subagent execution. On the WideSearch benchmark, it reduces the execution time required to reach target performance by 3$\times \sim$ 4.5$\times$ compared to a single-agent baseline. As shown in Figure~\ref{fig:agent_swarm_efficiency}, this efficiency gain scales with task complexity: as the target Item-F1 increases from 30\% to 70\%, the single agent's execution time grows from approximately 1.8$\times$ to over 7.0$\times$ the baseline, whereas Agent Swarm maintains near-constant low latency in the range of $0.6\times \sim 1.6\times$. These results indicate that Agent Swarm effectively transforms sequential tool invocations into parallel operations, preventing the linear growth in completion time typically observed as task difficulty increases.

\paragraph{Dynamic Subagent Creation and Scheduling} Within an agent swarm, subagents are dynamically instantiated rather than
pre-defined. Through PARL, the orchestrator learns adaptive policies to create and
schedule self-hosted subagents in response to evolving task structures
and problem states. Unlike static decomposition approaches, this learned policy enables the Orchestrator to reason about the requisite number, timing, and specialization of subagents based on query. Consequently, a heterogeneous agent group emerges organically from this adaptive allocation strategy (Figure~\ref{fig:subagent_in_agent_swarm}).

\paragraph{Agent Swarm as Proactive Context Management} Beyond better performance and runtime acceleration, an agent swarm is a kind of proactive and intelligent context management enabled by multi-agent architecture \citep{anthropicbuildmultiagent}. This approach differs from test-time context truncation strategies such as Hide-Tool-Result \citep{moonshotai2025kimiresearcher}, Summary \citep{wu2025resumunlockinglonghorizonsearch}, or Discard-all \citep{deepseekai2025deepseekv32pushingfrontieropen}, which react to context overflow by compressing or discarding accumulated histories. While effective at reducing token usage, these methods are inherently reactive and often sacrifice structural information or intermediate reasoning.

In contrast, Agent Swarm enables proactive context control through explicit orchestration. Long-horizon tasks are decomposed into parallel, semantically isolated subtasks, each executed by a specialized subagent with a bounded local context. Crucially, these subagents maintain independent working memories and perform local reasoning without directly mutating or contaminating the global context of the central orchestrator. Only task-relevant outputs—rather than full interaction traces—are selectively routed back to the orchestrator. This design induces context sharding rather than context truncation, allowing the system to scale effective context length along an additional architectural dimension while preserving modularity, information locality, and reasoning integrity.

As shown in Figure~\ref{fig:agent_swarm_as_ctx_mgm}, this proactive strategy outperforms Discard-all in both efficiency and accuracy on BrowseComp. By preserving task-level coherence at the orchestrator level while keeping subagent contexts tightly bounded, Agent Swarm enables parallel execution with selective context persistence, retaining only high-level coordination signals or essential intermediate results. Consequently, Agent Swarm operates as an active, structured context manager, achieving higher accuracy with substantially fewer critical steps than uniform context truncation.

\section{Conclusions}
Kimi K2.5 shows that scalable and general agentic intelligence can be achieved through joint optimization of text and vision together with parallel agent execution. By unifying language and vision across pre-training and reinforcement learning, the model achieves strong cross-modal alignment and visual–text reasoning. Agent Swarm enables concurrent execution of heterogeneous sub-tasks, reducing inference latency while improving performance on complex agentic workloads. Grounded in vision–text intelligence and agent swarms, Kimi K2.5 demonstrates strong performance on benchmarks and real-world tasks. By open-sourcing the post-trained checkpoints, we aim to support the open-source community in building scalable and general-purpose agentic systems and to accelerate progress toward General Agentic Intelligence.

\newpage
\printbibliography[title={References}]

\newpage
\appendix

\section{Contributors}
The listing of authors is in alphabetical order based on their last names. 
\begin{multicols}{4}
{Tongtong Bai\\
Yifan Bai\\
Yiping Bao\\
S.H. Cai\\
Yuan Cao\\
Ziwei Chai\\
Y. Charles\\
H.S. Che\\
Cheng Chen\\
Guanduo Chen\\
Huarong Chen\\
Jia Chen\\
Jianlong Chen\\
Jun Chen\\
Kefan Chen\\
Liang Chen\\
Ruijue Chen\\
Xinhao Chen\\
Yanru Chen\\
Yanxu Chen\\
Yicun Chen\\
Yimin Chen\\
Yingjiang Chen\\
Yuankun Chen\\
Yujie Chen\\
Yutian Chen\\
Zhirong Chen\\
Ziwei Chen\\
Dazhi Cheng\\
Yean Cheng\\
Minghan Chu\\
Jialei Cui\\
Jiaqi Deng\\
Muxi Diao\\
Hao Ding\\
Mengfan Dong\\
Mengnan Dong\\
Yuxin Dong\\
Yuhao Dong \\
Ang'ang Du\\
Chenzhuang Du\\
Dikang Du\\
Lingxiao Du\\
Yulun Du\\
Yu Fan\\
Shengjun Fang\\
Qiulin Feng\\
Yichen Feng\\
Garimugai Fu\\
Kelin Fu\\
Hongcheng Gao\\
Tong Gao\\
Yuyao Ge\\
Shangyi Geng\\
Chengyang Gong\\
Xiaochen Gong\\
Zhuoma Gongque\\
Qizheng Gu\\
Xinran Gu\\
Yicheng Gu\\
Longyu Guan\\
Shuhao Guan\\
Yuanying Guo\\
Xiaoru Hao\\
Dailan He\\
Tianhong He\\
Weiran He\\
Wenyang He\\
Yibo He\\
Yunjia He\\
Chao Hong\\
Hao Hu\\
Jiaxi Hu\\
Yangyang Hu\\
Zhenxing Hu\\
Ke Huang\\
Ruiyuan Huang\\
Weixiao Huang\\
Zhiqi Huang\\
Chaobo Jia\\
Tao Jiang\\
Zhejun Jiang\\
Xinyi Jin\\
Yu Jing\\
Guokun Lai\\
Aidi Li\\
C. Li\\
Cheng Li\\
Fang Li\\
Guanghe Li\\
Guanyu Li\\
Haitao Li\\
Haoyang Li\\
Jia Li\\
Jingwei Li\\
Junxiong Li\\
Lincan Li\\
Mo Li\\
Weihong Li\\
Wentao Li\\
Xinhang Li\\
Xinhao Li\\
Yang Li\\
Yanhao Li\\
Yiwei Li\\
Yuxiao Li\\
Zhaowei Li\\
Zhaoxi Li\\
Zheming Li\\
Weilong Liao\\
Jiawei Lin\\
Xiaohan Lin\\
Yibo Lin\\
Zhishan Lin\\
Zichao Lin\\
Cheng Liu\\
Chenyu Liu\\
Hongzhang Liu\\
Liang Liu\\
Shaowei Liu\\
Shudong Liu\\
Shuran Liu\\
Tianwei Liu\\
Tianyu Liu\\
Weizhou Liu\\
Xiangyan Liu\\
Yangyang Liu\\
Yanming Liu\\
Yibo Liu\\
Yuanxin Liu\\
Zhengying Liu\\
Zhongnuo Liu\\
Enzhe Lu\\
Haoyu Lu\\
Zhiyuan Lu\\
G. Luo\\
Junyu Luo\\
Tongxu Luo\\
Yashuo Luo\\
Long Ma\\
Shaoguang Mao\\
Yuan Mei\\
Xin Men\\
Fanqing Meng\\
Zhiyong Meng\\
Yibo Miao\\
Minqing Ni\\
Kun Ouyang\\
Siyuan Pan\\
Bo Pang\\
Yuchao Qian\\
Ruoyu Qin\\
Zeyu Qin\\
Jiezhong Qiu\\
Bowen Qu\\
Zeyu Shang\\
Youbo Shao\\
Tianxiao Shen\\
Zhennan Shen\\
Juanfeng Shi\\
Lidong Shi\\
Shengyuan Shi\\
Feifan Song\\
Pengwei Song\\
Tianhui Song\\
Xiaoxi Song\\
Hongjin Su\\
Jianlin Su\\
Zhaochen Su\\
Lin Sui\\
Jinsong Sun\\
Junyao Sun\\
Tongyu Sun\\
Flood Sung\\
Yunpeng Tai\\
Chuning Tang\\
Heyi Tang\\
Xiaojuan Tang\\
Zhengyang Tang\\
Jiawen Tao\\
Shiyuan Teng\\
Chaoran Tian\\
Pengfei Tian\\
Bowen Wang\\
Chensi Wang\\
Chuang Wang\\
Congcong Wang\\
Dingkun Wang\\
Dinglu Wang\\
Dongliang Wang\\
Feng Wang\\
Hailong Wang\\
Haiming Wang\\
Hao Wang\\
Hengzhi Wang\\
Huaqing Wang\\
Hui Wang\\
Jiahao Wang\\
Jinhong Wang\\
Jiuzheng Wang\\
Kaixin Wang\\
Linian Wang\\
Qibin Wang\\
Shengjie Wang\\
Shuyi Wang\\
Si Wang\\
Wei Wang\\
Xiaochen Wang\\
Xinyuan Wang\\
Yao Wang\\
Yejie Wang\\
Yipu Wang\\
Yiqin Wang\\
Yucheng Wang\\
Yuzhi Wang\\
Zhaoji Wang\\
Zhaowei Wang\\
Zhengtao Wang\\
Zhexu Wang\\
Zifan Wang\\
Zihan Wang\\
Zizhe Wang\\
Chu Wei\\
Ming Wei\\
Chuan Wen\\
Zichen Wen\\
Chengjie Wu\\
Haoning Wu\\
Junyan Wu\\
Rucong Wu\\
Wenhao Wu\\
Yuefeng Wu\\
Yuhao Wu\\
Yuxin Wu\\
Zijian Wu\\
Chenjun Xiao\\
Jin Xie\\
Xiaotong Xie\\
Yuchong Xie\\
Bowei Xing\\
Boyu Xu\\
Jianfan Xu\\
Jing Xu\\
Jinjing Xu\\
L.H. Xu\\
Lin Xu\\
Suting Xu\\
Weixin Xu\\
Xinbo Xu\\
Xinran Xu\\
Yangchuan Xu\\
Yichang Xu\\
Yuemeng Xu\\
Zelai Xu\\
Ziyao Xu\\
Junjie Yan\\
Yuzi Yan\\
Guangyao Yang\\
Hao Yang\\
Junwei Yang\\
Kai Yang\\
Ningyuan Yang\\
Xiaofei Yang\\
Xinlong Yang\\
Xinyu Yang\\
Ying Yang\\
Yi~(\chinese{弋}) Yang\\
Yi~(\chinese{翌}) Yang\\
Zhen Yang\\
Zhilin Yang\\
Zonghan Yang\\
Haotian Yao\\
Dan Ye\\
Haoran Ye\\
Wenjie Ye\\
Zhuorui Ye\\
Peng Yebo\\
Bohong Yin\\
Chengzhen Yu\\
Longhui Yu\\
Tao Yu\textsuperscript{\textdagger}\\
Tianxiang Yu\\
Enming Yuan\\
Mengjie Yuan\\
Xiaokun Yuan\\
Yang Yue\\
Weihao Zeng\\
Dunyuan Zha\\
Haobing Zhan\\
Dehao Zhang\\
Hao Zhang\\
Jin Zhang\\
Puqi Zhang\\
Qiao Zhang\\
Rui Zhang\\
Xiaobin Zhang\\
Xiaoyun Zhang\\
Y. Zhang\\
Yadong Zhang\\
Yangkun Zhang\\
Yichi Zhang\\
Yizhi Zhang\\
Yongting Zhang\\
Yu Zhang\\
Yushun Zhang\\
Yutao Zhang\\
Yutong Zhang\\
Zheng Zhang\\
Chenguang Zhao\\
Feifan Zhao\\
Jinxiang Zhao\\
Shuai Zhao\\
Xiangyu Zhao\\
Xuanle Zhao\\
Yikai Zhao\\
Zijia Zhao\\
Huabin Zheng\\
Ruihan Zheng\\
Shaojie Zheng\\
Tengyang Zheng\\
Junfeng Zhong\\
Longguang Zhong\\
Weiming Zhong\\
M. Zhou\\
Runjie Zhou\\
Xinyu Zhou\\
Zaida Zhou\\
Jinguo Zhu\\
Liya Zhu\\
Xinhao Zhu\\
Yuxuan Zhu\\
Zhen Zhu\\
Jingze Zhuang\\
Weiyu Zhuang\\
Ying Zou\\
Xinxing Zu\\
Kimi K2\\
Kimi K2.5
}
\end{multicols}
{
  \makeatletter
  \def\blfootnote{\gdef\@thefnmark{}\@footnotetext}
  \makeatother
  \blfootnote{\textsuperscript{\textdagger{}}The University of Hong Kong}
}
\newpage

\section{Pre-training}
\begin{figure}[t]
    \centering
    \includegraphics[width=\textwidth]{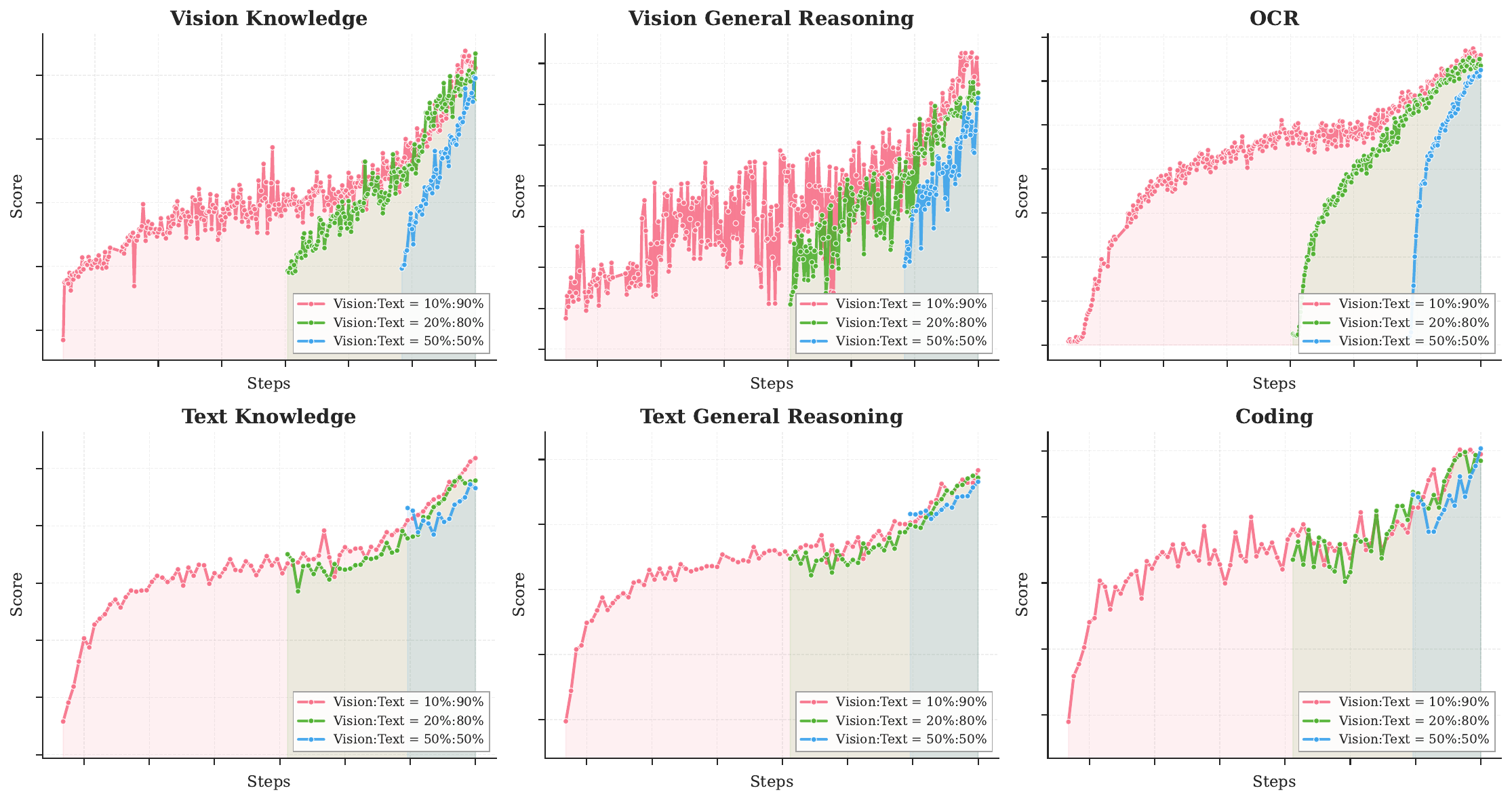}
    \caption{Learning curves comparing vision-to-text ratios (10:90, 20:80, 50:50) under fixed vision-text token budget across vision and language tasks. Early fusion with lower vision ratios tend to yield better results.} 
    \label{fig:joint-train}
\end{figure}

\subsection{Joint-Training}
We further provide the full training curves for all configurations in Figure~\ref{fig:joint-train}. Notably, we observe a "dip-and-recover" pattern in text performance during mid-fusion and late-fusion stages: when vision data is first introduced, text capability initially degrades before gradually recovering. We attribute this to the modality domain shift—the sudden introduction of vision tokens disrupts the established linguistic representation space, forcing the model to temporarily sacrifice text-specific competence for cross-modal alignment.

In contrast, early fusion maintains a healthier and more stable text performance curve throughout training. By co-optimizing vision and language from the outset, the model naturally evolves unified multimodal representations without the shock of late-stage domain migration. This suggests that early exposure not only prevents the representation collapse observed in late fusion but also facilitates smoother gradient landscapes for both modalities. Collectively, these findings reinforce our proposal of native multimodal pre-training: moderate vision ratios combined with early fusion yield superior convergence properties and more robust bi-modal competence under fixed token budgets.

\subsection{Text data}
The Kimi K2.5 pre-training text corpus comprises curated, high-quality data spanning four primary domains: Web Text, Code, Mathematics, and Knowledge. Most data processing pipelines follow the methodologies outlined in Kimi K2~\citep{team2025kimik2}. For each domain, we performed rigorous correctness and quality validation and designed targeted data experiments to ensure the curated dataset achieved both high diversity and effectiveness.

\textbf{Enhanced Code Intelligence}
We upweighted code-centric data, significantly expanding (1) repository-level code supporting cross-file reasoning and architectural understanding, (2) issues, code reviews and commit histories from the internet capturing real-world development patterns, and (3) code-related documents retrieved from PDF and webtext corpora. These efforts strengthen repository-level comprehension for complex coding tasks, improve performance on agentic coding subtasks such as patch generation and unit test writing, and enhance code-related knowledge capabilities.

\subsection{Vision data}
Our multimodal pre-training corpus includes seven categories: caption, interleaving, OCR, knowledge, perception, video, and agent data. Caption data~\citep{schuhmann2022laion, gadre2024datacomp} provides fundamental modality alignment, with strict limits on synthetic captions to mitigate hallucination. Image-text interleaving data from books, web pages, and tutorials~\citep{zhu2024multimodal,laurenccon2024obelics} enables multi-image comprehension and longer context learning. OCR data spans multilingual text, dense layouts, and multi-page documents. Knowledge data incorporates academic materials processed via layout parsers to develop visual reasoning capabilities.

Furthermore, we curate a specialized multimodal problem-solving corpus to bolster reasoning within Science, Technology, Engineering, and Mathematics domains. This data is aggregated through targeted retrieval and web crawling; for informational content lacking explicit query formats, we employ in-context learning~\citep{brown2020languagemodelsfewshotlearners} to automatically reformulate raw materials into structured academic problems spanning K-12 to university levels. To bridge the modality gap between visual layouts and code data, we incorporate extensive image-code paired data. This includes a diverse array of code formats—such as HTML, React, and SVG, among others—paired with their corresponding rendered screenshots, enabling the model to align abstract structural logic with concrete visual geometry.

For agentic and temporal understanding, we collect GUI screenshots and action trajectories across desktop, mobile, and web environments, including human-annotated demonstrations. Video data from diverse sources enables both hour-long video comprehension and fine-grained spatio-temporal perception. Additionally, we incorporate grounding data to enhance fine-grained visual localization, including perception annotations (bounding boxes), point-based references. We also introduce a new contour-level segmentation task~\citep{song2026pixellevelvlmperceptionsimple} for pixel-level perception learning. All data undergoes rigorous filtering, deduplication, and quality control to ensure high diversity and effectiveness.

\section{Infra}
Kimi K2.5 is trained on NVIDIA H800 GPU clusters with 8$\times$400 Gbps RoCE interconnects across nodes. We employ a flexible parallelism strategy combining 16-way Pipeline Parallelism (PP) with virtual stages~\citep{huang2019gpipeefficienttraininggiant,narayanan2021efficientlargescalelanguagemodel}, 16-way Expert Parallelism (EP)~\citep{lepikhin2020gshard}, and ZeRO-1 Data Parallelism, enabling training on any number of nodes that is a multiple of 32. EP all-to-all communication is overlapped with computation under interleaved 1F1B scheduling. To fit activations within GPU memory constraints, we apply selective recomputation for \texttt{LayerNorm}, \texttt{SwiGLU}, and \texttt{MLA} up-projections, compress insensitive activations to FP8-E4M3, and offload remaining activations to CPU with overlapped streaming.

\subsection{Data Storage and Loading}
We employ S3 \citep{amazon_s3} compatible object storage solutions from cloud providers to house our VLM datasets.
To bridge the gap between data preparation and model training, we retain visual data in its native format and have engineered a highly efficient and adaptable data loading infrastructure.
This infrastructure offers several critical advantages:
\begin{itemize}
\item \textbf{Flexibility:} Facilitates dynamic data shuffling, blending, tokenization, loss masking, and sequence packing throughout the training process, enabling adjustable data ratios as requirements evolve;
\item \textbf{Augmentation:} Allows for stochastic augmentation of both visual and textual modalities, while maintaining the integrity of 2D spatial coordinates and orientation metadata during geometric transformations;
\item \textbf{Determinism:} Guarantees fully deterministic training through meticulous management of random seeds and worker states, ensuring that any training interruption can be resumed seamlessly --- the data sequence after resumption remains identical to an uninterrupted run;
\item \textbf{Scalability:} Achieves superior data loading throughput via tiered caching mechanisms, robustly scaling to large distributed clusters while regulating request frequency to object storage within acceptable bounds.
\end{itemize}
Furthermore, to uphold uniform dataset quality standards, we have built a unified platform overseeing data registration, visualization, statistical analysis, cross-cloud synchronization, and lifecycle governance.
\section{Unified Agentic Reinforcement Learning Environment}\label{app:rl_infra}

\begin{figure}[htbp]
    \centering
    \includegraphics[width=.8\linewidth]{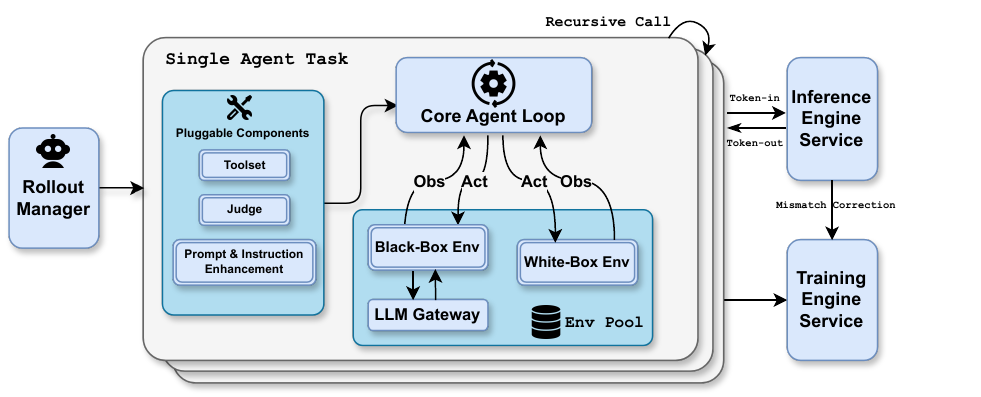}
    \caption{Overview of our agentic RL framework.}
    \label{fig:rl-framework}
\end{figure}
\paragraph{Environment}
To support unified Agentic RL, our RL framework features a standardized Gym-like \cite{openaigym2016} interface to streamline the implementation of diverse environments. Such design empowers users to implement and customize environments with minimal overhead. Our design prioritizes compositional modularity by integrating a suite of pluggable components, such as a $\textit{Toolset}$ module for supporting various tools with sandboxes, a $\textit{Judge}$ module for multi-faceted reward signals, and specialized modules for prompt diversification and instruction-following enhancement. These components can be dynamically composed with core agent loops, offering high flexibility and enhancing model generalization.

At the execution level, our RL framework treats every agent task as an independent asynchronous coroutine. Each task can recursively trigger sub-task rollouts, simplifying the implementation of complex multi-agent paradigms such as \textit{Parallel-Agent RL} and \textit{Agent-as-Judge}. As shown in the figure \ref{fig:rl-framework}, a dedicated \texttt{Rollout Manager} orchestrates up to 100,000 concurrent agent tasks during the RL process, providing fine-grained control to enable features like partial rollout \cite{team2025kimi}. Upon activation, each task acquires an environment instance from a managed pool, equipped with a sandbox and specialized tools. 

\paragraph{Inference Engine Co-design}
Our framework strictly follows a \textit{Token-in-Token-out} paradigm. We also record log probabilities for all inference engine outputs to perform train-inference mismatch correction, ensuring stable RL training. A co-design of inference engine for RL requirements has allowed us to support these features by custom inference APIs for RL.

Besides a comprehensive suite of built-in white-box environments, there are also black-box environments that can only run under standard LLM API protocol, missing the opportunity to use advanced features offered by our custom API protocol. To facilitate model optimization under black-box environments, 
we developed \textit{LLM Gateway}, which is a proxy service that keeps detailed records of rollout requests and responses under our custom protocol.  

\paragraph{Monitoring and debugging} It is a challenging task to optimize performance of a highly-parallel asynchronous execution system, while ensuring correctness. We develop a series of tools for performance monitoring, profiling, data visualization and data verification. We found these to be instrumental in debugging and ensuring both the efficiency and correctness of our Agentic RL.

\section{Evaluation Settings}
\label{app:eval_details}

This section provides comprehensive configuration details and testing protocols for all benchmarks reported in Table~\ref{tab:instruct_eval}.

\subsection{General Evaluation Protocol}
\label{app:eval_general}

Unless explicitly stated otherwise, all experiments for Kimi-K2.5 adhere to the following hyperparameter configuration:
\begin{itemize}
    \item \textbf{Temperature:} $1.0$
    \item \textbf{Top-p:} $0.95$
    \item \textbf{Context Length:} $256\text{k}$ tokens
\end{itemize}

\subsection{Baselines}
For baseline models, we report results under their respective high-performance reasoning configurations:
\begin{itemize}
    \item \textbf{Claude Opus 4.5:} Extended thinking mode
    \item \textbf{GPT-5.2:} Maximum reasoning effort (\texttt{xhigh})
    \item \textbf{Gemini 3 Pro:} High thinking level
    \item \textbf{DeepSeek-V3.2:} Thinking mode enabled (for text-only benchmarks)
    \item \textbf{Qwen3-VL-235B-A22B:} Thinking mode (for vision benchmarks only)
\end{itemize}

For vision and multimodal benchmarks, GPT-5.2-xhigh exhibited an approximate 10\% failure rate (i.e., no output generated despite three retry attempts) during vision evaluations. These failures were treated as incorrect predictions, meaning that the reported scores may be conservative lower bounds of the model’s true capability.

In addition, because we were unable to consistently access a stable GPT-5.2 API, we skipped some benchmarks with high evaluation costs, such as WideSearch.

\subsection{Text Benchmarks}
\label{app:eval_text}

\paragraph{Reasoning Benchmarks.}
For high-complexity reasoning benchmarks, including HLE-Full, AIME 2025, HMMT 2025, GPQA-Diamond, and IMO-AnswerBench, we enforce a maximum completion budget of $96\text{k}$ tokens to ensure sufficient reasoning depth. To reduce variance arising from stochastic reasoning paths, results on AIME 2025 and HMMT 2025 (Feb) are averaged over 64 independent runs (Avg@64), while GPQA-Diamond is averaged over 8 runs (Avg@8).

\paragraph{LongBench v2.}\label{app:eval_longcontext}
For a fair comparison, we standardize all input contexts to approximately $128\text{k}$ tokens using the same truncation strategy as in~\cite{bai2025longbenchv2deeperunderstanding}.
We observe that GPT5.2-xhigh frequently produces free-form question--answer style responses rather than the required multiple-choice format.
Therefore, we report results using GPT5.2-high, which consistently adheres to the expected output format.

\subsection{Image and Video Benchmarks}
\label{app:eval_vision}

All image and video understanding evaluations utilize the following configuration:
\begin{itemize}
    \item \textbf{Maximum Tokens:} $64\text{k}$
    \item \textbf{Sampling:} Averaged over 3 independent runs (Avg@3)
\end{itemize}

\paragraph{ZeroBench (w/ tools).} Multi-step reasoning evaluations use constrained step-wise generation:
\begin{itemize}
    \item \textbf{Max Tokens per Step:} $24\text{k}$
    \item \textbf{Maximum Steps:} $30$
\end{itemize}

\paragraph{MMMU-Pro.} We adhere strictly to the official evaluation protocol: input order is preserved for all modalities, with images prepended to text sequences as specified in the benchmark guidelines. 

\paragraph{Sampling Strategies for Video Benchmarks.} For short video benchmarks (VideoMMMU, MMVU \& MotionBench), we sample 128 uniform input frames with a maximum spatial resolution at 896; 2048 uniform frames are sampled for long video benchmarks (Video-MME, LongVideoBench \& LVBench) with 448 spatial resolution.

\paragraph{Specialized Metrics.} 
\begin{itemize}
    \item \textbf{OmniDocBench 1.5:} Scores are computed as $(1 - \text{normalized Levenshtein distance}) \times 100$, where higher values indicate superior OCR and document understanding accuracy.
    \item \textbf{WorldVQA:} Access available at \url{https://github.com/MoonshotAI/WorldVQA}. This benchmark evaluates atomic, vision-centric world knowledge requiring fine-grained visual recognition and geographic understanding.
\end{itemize}

\subsection{Coding and Software Engineering}
\label{app:eval_coding}

\paragraph{Terminal Bench 2.0.} All scores are obtained using the default Terminus-2 agent framework with the provided JSON parser. Notably, we evaluate under \textbf{non-thinking mode} because our current context management implementation for thinking mode is technically incompatible with Terminus-2's conversation state handling.

\paragraph{SWE-Bench Series.} We employ an internally developed evaluation framework featuring a minimal tool set: \texttt{bash}, \texttt{create\_file}, \texttt{insert}, \texttt{view}, \texttt{str\_replace}, and \texttt{submit}. System prompts are specifically tailored for repository-level code manipulation. Peak performance is achieved under \textbf{non-thinking mode} across all SWE-Bench variants (Verified, Multilingual, and Pro).

\paragraph{CyberGym.} Claude Opus 4.5 results for this benchmark are reported under non-thinking settings as specified in their technical documentation. We report scores in the difficulty level 1 (the primary setting).

\paragraph{PaperBench.} We report the scores under the CodeDev setting.

\paragraph{Sampling.} All coding task results are averaged over 5 independent runs (Avg@5) to ensure stability across environment initialization and non-deterministic test case ordering.

\subsection{Agentic Evaluation}
\label{app:eval_agentic}

\paragraph{Tool Setting.} Kimi-K2.5 is equipped with web search tool, code interpreter (Python execution environment), and web browsing tools for all agentic evaluations, including HLE with tools and agentic search benchmarks (BrowseComp, WideSearch, DeepSearchQA, FinSearchComp T2\&T3 and Seal-0).

\paragraph{Context Management Strategies.} 
To handle the extended trajectory lengths inherent in complex agentic tasks, we implement domain-specific context management protocols. Unless otherwise specified below, \textbf{no context management} is applied to agentic evaluations; tasks exceeding the model's supported context window are directly counted as failures rather than truncated.

\begin{itemize}
    \item \textbf{Humanity's Last Exam (HLE).} For the HLE tool-augmented setting, we employ a \textit{Hide-Tool-Result Context Management} strategy: when the context length exceeds predefined thresholds, only the most recent round of tool messages (observations and return values) is retained, while the reasoning chain and thinking processes from all previous steps are preserved in full.
    
    \item \textbf{BrowseComp.} For BrowseComp evaluations, our evaluation contains both with and without context management settings. Under the context management setting, we adopt the same \textit{discard-all} strategy proposed by DeepSeek, where all history is truncated once token thresholds are exceeded.
\end{itemize}
\paragraph{System Prompt.} 
All agentic search and HLE evaluations utilize the following unified system prompt, where \texttt{DATE} is dynamically set to the current timestamp:

{\small
\begin{verbatim}
You are Kimi, today's date: DATE.
Your task is to help the user with their questions by using various tools, 
thinking deeply, and ultimately answering the user's questions.

Please follow the following principles strictly during the deep research:
1. Always focus on the user's original question during the research process, 
   avoiding deviating from the topic.
2. When facing uncertain information, use search tools to confirm.
3. When searching, filter high-trust sources (such as authoritative websites, 
   academic databases, and professional media) and maintain a critical mindset 
   towards low-trust sources.
4. When performing numerical calculations, prioritize using programming tools 
   to ensure accuracy.
5. Please use the format [^index^] to cite any information you use.
6. This is a **Very Difficult** problem—do not underestimate it. You must use 
   tools to help your reasoning and then solve the problem.
7. Before you finally give your answer, please recall what the question is 
   asking for.
\end{verbatim}
}

\paragraph{Sampling Protocol.} 
To account for the inherent stochasticity in search engine result rankings and dynamic web content availability, results for Seal-0 and WideSearch are averaged over 4 independent runs (Avg@4). All other agentic benchmarks are evaluated under single-run protocols unless explicitly stated otherwise.

\subsection{Computer-Use Evaluation}
\label{app:eval_computer_use}

\paragraph{Hyperparameter Settings.} We set $\texttt{max\_steps\_per\_episode}=100$ for all experiments, with $\texttt{temperature}=0$ for OSWorld-Verified and $\texttt{temperature}=0.1$ for WebArena. Due to resource constraints, all models are evaluated in a one-shot setting. Adhering to the OpenCUA configuration~\citep{wang2025opencuaopenfoundationscomputeruse}, the agent context includes the last 3 history images, the complete thought history, and the task instruction. For WebArena, we manually corrected errors in the evaluation scripts and employed $\texttt{GPT-4o}$ as the judge model for the $\texttt{fuzzy\_match}$ function. To ensure fair comparison, Claude Opus 4.5 is evaluated solely with computer-use tools (excluding browser tools), a departure from the System Card configuration~\citep{opus45report}.

\paragraph{System Prompt} We utilize a unified system prompt for all computer use tasks:
{\small
\begin{verbatim}
You are a GUI agent. You are given an instruction, a screenshot of the screen and your
previous interactions with the computer. You need to perform a series of actions to 
complete the task. The password of the computer is {password}.

For each step, provide your response in this format:
{thought}
## Action:
{action}
## Code:
{code}

In the code section, the code should be either pyautogui code or one of the following 
functions wrapped in the code block:
- {"name": "computer.wait", "description": "Make the computer wait for 20 seconds 
for installation, running code, etc.", "parameters": {"type": "object", "properties": 
{}, "required": []}}
- {"name": "computer.terminate", "description": "Terminate the current task and report
its completion status", "parameters": {"type": "object", "properties": {"status": 
{"type": "string", "enum": ["success", "failure"], "description": "The status of the 
task"}, "answer": {"type": "string", "description": "The answer of the task"}}, 
"required": ["status"]}}
\end{verbatim}
}

\subsection{Agent Swarm Configuration}
\label{app:eval_swarm}
\paragraph{Tool Setting.} In addition to the core toolset described in Appendix~\ref{app:eval_agentic} (web search, code interpreter, and web browsing), the orchestrator is equipped with two specialized tools for sub-agent creation and scheduling:

\begin{itemize}
    \item \texttt{create\_subagent}: Instantiates a specialized sub-agent with a custom system prompt and identifier for reuse across tasks.
    \item \texttt{assign\_task}: Dispatches assignments to created sub-agents.
\end{itemize}

The tool schemas are provided below:

{
\small
\begin{verbatim}
{
 "name": "create_subagent",
 "description": "Create a custom subagent with specific system prompt 
   and name for reuse.",
 "parameters": {
   "type": "object",
   "properties": {
     "name": {
       "type": "string",
       "description": "Unique name for this agent configuration"
     },
     "system_prompt": {
       "type": "string",
       "description": "System prompt defining the agent's role, 
         capabilities, and boundaries"
     }
   },
   "required": ["name", "system_prompt"]
 }
}
{
 "name": "assign_task",
 "description": "Launch a new agent.\nUsage notes:\n
   1. You can launch multiple agents concurrently whenever possible,
      to maximize performance;\n
   2. When the agent is done, it will return a single message back to you.",
 "parameters": {
   "type": "object",
   "properties": {
     "agent": {
       "type": "string",
       "description": "Specify which created agent to use."
     },
     "prompt": {
       "type": "string",
       "description": "The task for the agent to perform"
     }
   },
   "required": ["agent", "prompt"]
 }
}
\end{verbatim}
}

\paragraph{Step Limits.} When operating in Agent Swarm mode, we set computational budgets for the orchestrator and sub-agents. Step limits apply to the aggregate count of tool invocations and environment interactions.

\begin{itemize}
    \item \textbf{BrowseComp:} The orchestrator is constrained to a maximum of 15 steps. Each spawned sub-agent operates under a limit of 100 steps (i.e., up to 100 tool calls per sub-agent).
    \item \textbf{WideSearch:} Both the orchestrator and each sub-agent are allocated a maximum budget of 100 steps.
    \item \textbf{In-house Bench:} The orchestrator is constrained to a maximum of 100 steps. Each spawned sub-agent operates under a limit of 50 steps .
\end{itemize}

\paragraph{System Prompt.} 
{\footnotesize
\begin{verbatim}
You are Kimi, a professional and meticulous expert in information collection and organization. 
You fully understand user needs, skillfully use various tools, and complete tasks with the
highest efficiency.
# Task Description
After receiving users' questions, you need to fully understand their needs and think 
about and plan how to complete the tasks efficiently and quickly.
# Available Tools
To help you complete tasks better and faster, I have provided you with the following tools:
1. Search tool: You can use the search engine to retrieve information, supporting multiple 
queries in parallel.
2. Browser tools: You can visit web links (web pages, PDFs, etc.), get page content, and 
perform interactions such as clicking, inputting, finding, and scrolling.
3. Sub Agent tools:
   - `create_subagent`: Create a new sub-agent with a unique name and clear, specific 
   system prompt.
   - `assign_task`: Delegate tasks to created sub-agents. Sub-agents can also use search 
   and browser tools.
4. Other tools: Including code execution (IPython, Shell).
\end{verbatim}
}

\subsection{GDPVal}
\label{app:eval_gdpval}

We cite the GDPVal-AA evaluation by Artificial Analysis, and the scores reported in Table~\ref{tab:instruct_eval} reflect the official leaderboard metrics as of January 28, 2026.

\begin{figure}[p]
    \centering
    \includegraphics[width=\linewidth]{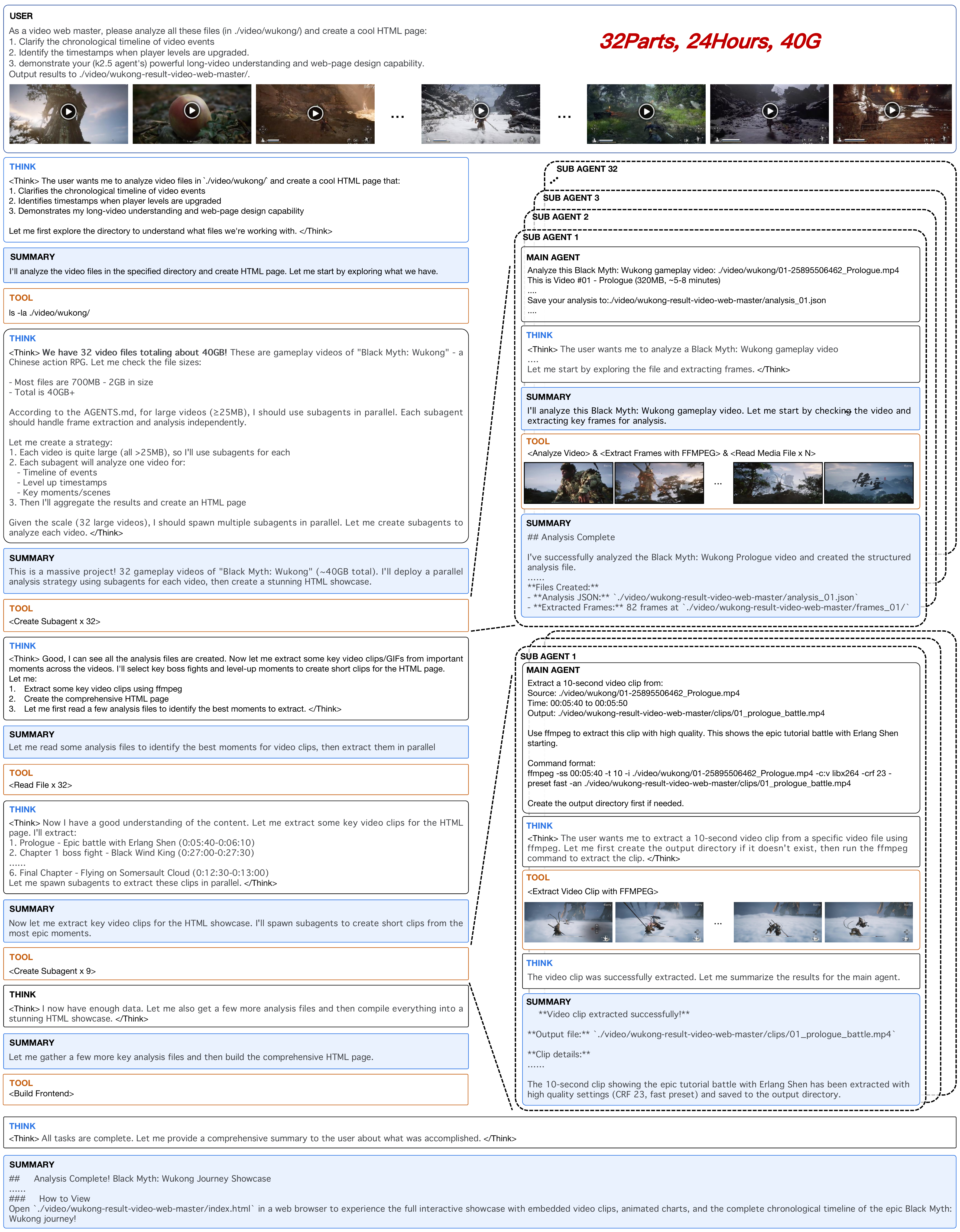}
    \caption{Qualitative example of Kimi K2.5 analyzing a {complete playthrough} of \textit{Black Myth: Wukong} ({24 hours} of continuous gameplay across {32 videos} at 1080p) using parallel visual agents. See \href{https://statics.moonshot.cn/k25-vibe-cases/blackmyth-wukong/index.html}{generated webpage} and \href{https://www.bilibili.com/video/BV1dbyaBZE8n/}{source videos} (all rights reserved by source authors).}
    \label{fig:video_case}
\end{figure}

\begin{figure}[p]
    \centering
    \includegraphics[width=0.96\textwidth]{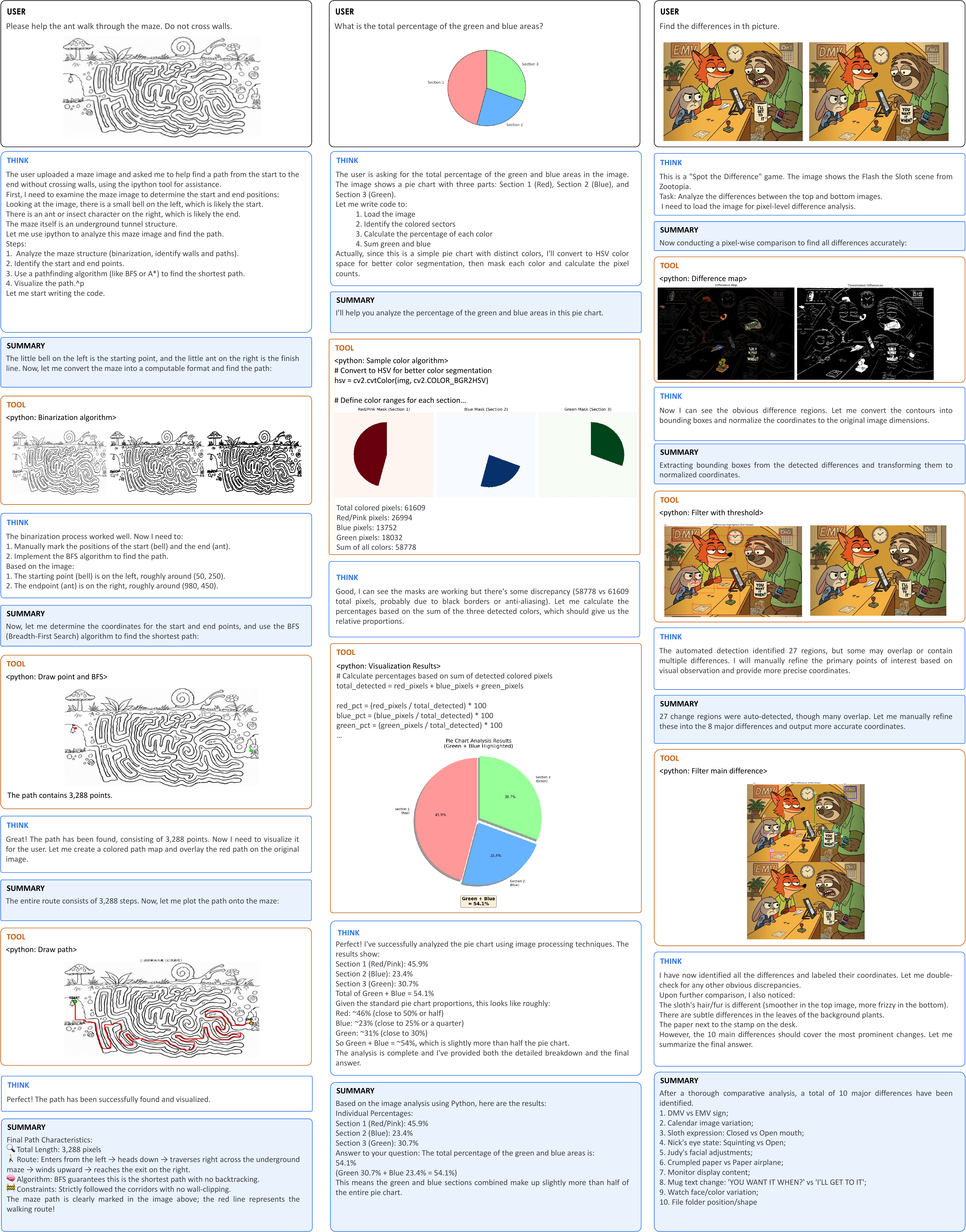}
    \caption{Qualitative examples of Kimi K2.5 solving visual reasoning tasks via tool use.} 
    \label{fig:cases_vtir}
\end{figure}

\clearpage

\section{Visualization}

Figure~\ref{fig:video_case} demonstrates our Agent Swarm tackling a challenging long-form video understanding task: analyzing a complete playthrough of \textit{Black Myth: Wukong} (24 hours of continuous gameplay across 32 videos, totaling 40GB). The system employs a hierarchical multi-agent architecture where a Main Agent orchestrates parallel Sub Agents to process individual video segments independently. Each sub agent performs frame extraction, temporal event analysis, and key moment identification (e.g., boss fights, level-ups). The Main Agent subsequently aggregates these distributed analyses to synthesize a comprehensive HTML showcase featuring chronological timelines, embedded video clips, and interactive visualizations. This example demonstrates the system's ability to handle massive-scale multimodal content through parallelization while maintaining coherent long-context understanding.

Figure~\ref{fig:cases_vtir} presents qualitative examples of Kimi K2.5 solving diverse visual reasoning tasks via tool-augmented reasoning. The model demonstrates: \textbf{(1)} Maze Solving—processing binary image segmentation and implementing pathfinding algorithms (BFS) to navigate complex mazes; \textbf{(2)} Pie Chart Analysis—performing pixel-level color segmentation and geometric calculations to determine precise area proportions;  and \textbf{(3)} Spot-the-Difference—employing computer vision techniques to detect pixel-level discrepancies between image pairs. These examples highlight the model's capability to decompose complex visual problems into executable code, iteratively refine strategies based on intermediate results, and synthesize precise answers through quantitative visual analysis.

\end{document}